\documentclass[preprint,12pt,authoryear]{elsarticle}

\usepackage{amssymb}
\usepackage{amsmath}
\usepackage{cite}
\usepackage{xcolor}
\definecolor{orange}{rgb}{1,0.5,0}  % RGB模型下的橙色

\usepackage{subfig}
\usepackage{multirow}
\usepackage{makecell}

\journal{Robotics and Autonomous Systems}

\begin{document}

\begin{frontmatter}



\title{Unified Vertex Motion Estimation for Integrated Video Stabilization and Stitching in Tractor-Trailer Wheeled Robots} %% Article title


% \author{} %% Author name

% \affiliation{organization={},%Department and Organization
% %% Title and authors
% \title{Your Title Here}

%% Authors and affiliations
\author[bit]{Hao Liang}
\author[bit]{Zhipeng Dong}
\author[bit]{Hao Li}
\author[bit]{Yufeng Yue}
\author[bit,nust]{Mengyin Fu}
\author[bit]{Yi Yang\corref{cor1}}
\ead{yang_yi@bit.edu.cn}

%% Corresponding author information
\cortext[cor1]{Corresponding author}

%% Affiliations
\affiliation[bit]{organization={School of Automation, Beijing Institute of Technology}, 
            city={Beijing},
            postcode={100081}, 
            country={China}}

\affiliation[nust]{organization={School of Automation, Nanjing University of Science and Technology}, 
            city={Nanjing},
            postcode={210094}, 
            country={China}}

%% Abstract
\begin{abstract}
% 拖挂车被期望在园区物流、干线运输等领域执行全面的感知任务。拖挂车感知主要面临三大挑战，非刚性连接的车头与车身带来的位置变化，车头运动震动与车身运动震动不同步，大尺寸车身带来的相机极大视差。In this paper, a novel unified video stabilization and stitching framework is proposed in unknown environments. The key novelty of this work is the proposing of a unified framework to formulate the video stabilization and stitching problem with a Unified Vertex Motion model, as well as its realization in real tractor-trailer vehicles. 为了建立stabilization和stitching之间的关系，一种新的Unified Vertex Motion被提出。这种motion统一时序motion，即针对stabilization，与空间motion，即针对stitching。考虑到stablization与stitching对优化函数的异构性，提出一种加权的代价函数方法，可以针对stabilization或stitching进行调节。提出的Unified Vertex Motion Video Stabilization and Stitching方法在各种具有挑战性的场景中进行了验证，证明了其在trator-trailer实际任务中的准确性和实用性。
Tractor-trailer wheeled robots need to perform comprehensive perception tasks to enhance their operations in areas such as logistics parks and long-haul transportation. The perception of these robots \textcolor{black}{faces} three major challenges: the asynchronous vibrations between the tractor and trailer, the relative pose change between the tractor and trailer, and the significant camera parallax caused by the large size. \textcolor{black}{In this paper, we employ the Dual Independence Stabilization Motion Field Estimation method to address asynchronous vibrations between the tractor and trailer, effectively eliminating conflicting motion estimations for the same object in overlapping regions. We utilize the Random Plane-based Stitching Motion Field Estimation method to tackle the continuous relative pose changes caused by the articulated hitch between the tractor and trailer, thus eliminating dynamic misalignment in overlapping regions. Furthermore, we apply the Unified Vertex Motion Estimation method to manage the challenges posed by the tractor-trailer's large physical size, which results in severely low overlapping regions between the tractor and trailer views, thus preventing distortions in overlapping regions from exponentially propagating into non-overlapping areas.} Furthermore, this framework has been successfully implemented in real tractor-trailer wheeled robots.  The proposed Unified Vertex Motion Video Stabilization and Stitching method has been thoroughly tested in various challenging scenarios, demonstrating its accuracy and practicality in real-world robot tasks. The code is available at https://github.com/lhlawrence/UVM-VSS.
\end{abstract}

\begin{graphicalabstract}
\includegraphics[width=\linewidth]{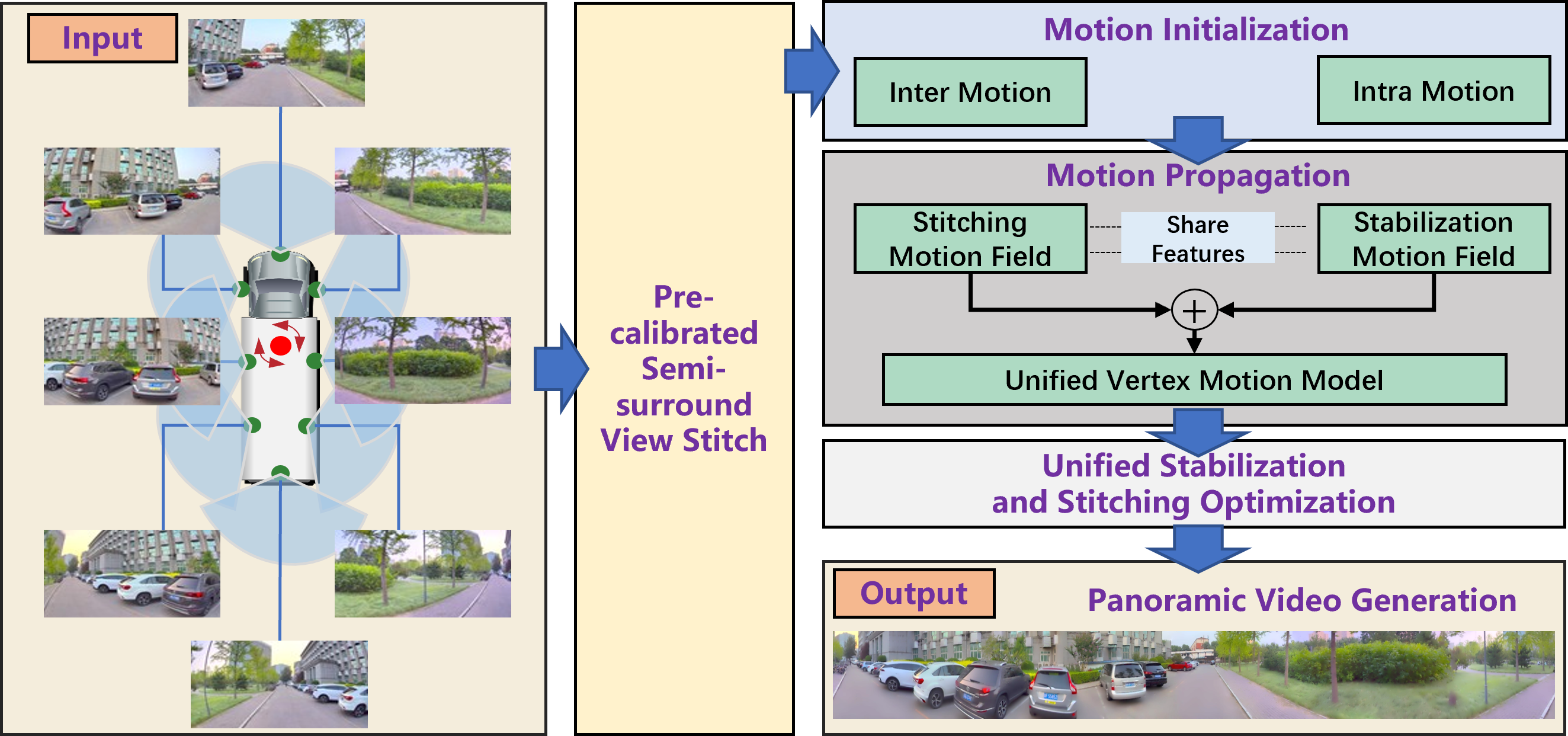}
\end{graphicalabstract}

\begin{highlights}
\item \textcolor{black}{We design a Dual Independence Stabilization Motion Fields Estimation method that independently characterizes motion patterns of the tractor and trailer units.}
\item \textcolor{black}{We propose a Random Plane based Stitching Motion Estimation method, which directly derives pixel-level displacements from feature motions without relying on pose estimation or geometric priors. }
\item \textcolor{black}{We develop a Unified Vertex Motion Estimation method, which estimates spatiotemporal motions, including those in non-overlapping regions. The corresponding joint optimization framework accounting for combined stitching-stabilization effects solves the problem of distortion propagation through low-overlapping regions that cause catastrophic geometric distortion amplification.}
\item We have implemented a practical hardware and software solution for a surround-view system that can be deployed on \textcolor{black}{tractor-trailer} wheeled robots.
\end{highlights}

%% Keywords

\end{frontmatter}



\section{Introduction}
\label{intro}
%轮式机器人在人类的日常生活中起到了越来越重要的作用。在轮式机器人中，有一种特殊的类型叫 tractor-trailer wheeled robot(TTWR)。 Typically, it consists of two wheeled mobile robots connected by a linkage mechanism. 其由具有动力装置的trator负责运动，具有负载能力的trailer负责运载。现有针对TTWR的研究主要集中在系统建模、路径规划、运动控制方面，而作为机器人系统中重要的感知单元，并针对tractor-trailer wheeled robot方面的研究急需得到发展。环视拼接与防抖是其中比较基础又重要的前端工作，给后面的技术框架提供无盲区的环视图像预处理数据。
Wheeled robots are playing an increasingly significant role in our daily lives. Among these, a specialized type known as the tractor-trailer wheeled robot (TTWR) stands out\citep{TTWR}. Typically, a TTWR consists of two wheeled mobile robots connected by a linkage mechanism\citep{miao}. The tractor, equipped with a propulsion system, is responsible for movement, while the trailer, designed for load-carrying, handles transportation. Current research on TTWRs primarily focuses on path planning\citep{TTWR}\citep{plan1}\citep{plan2} and motion control\citep{control1}\citep{control2}\citep{control3}. However, as a crucial component of the robotic system, research specifically targeting the perception capabilities of TTWRs is urgently needed. Surround-view perception is \textcolor{black}{a fundamental task} in this area, providing seamless panoramic image preprocessing data for subsequent technological frameworks.

\begin{figure}[!t]
\centering
\includegraphics[width=0.5\linewidth]{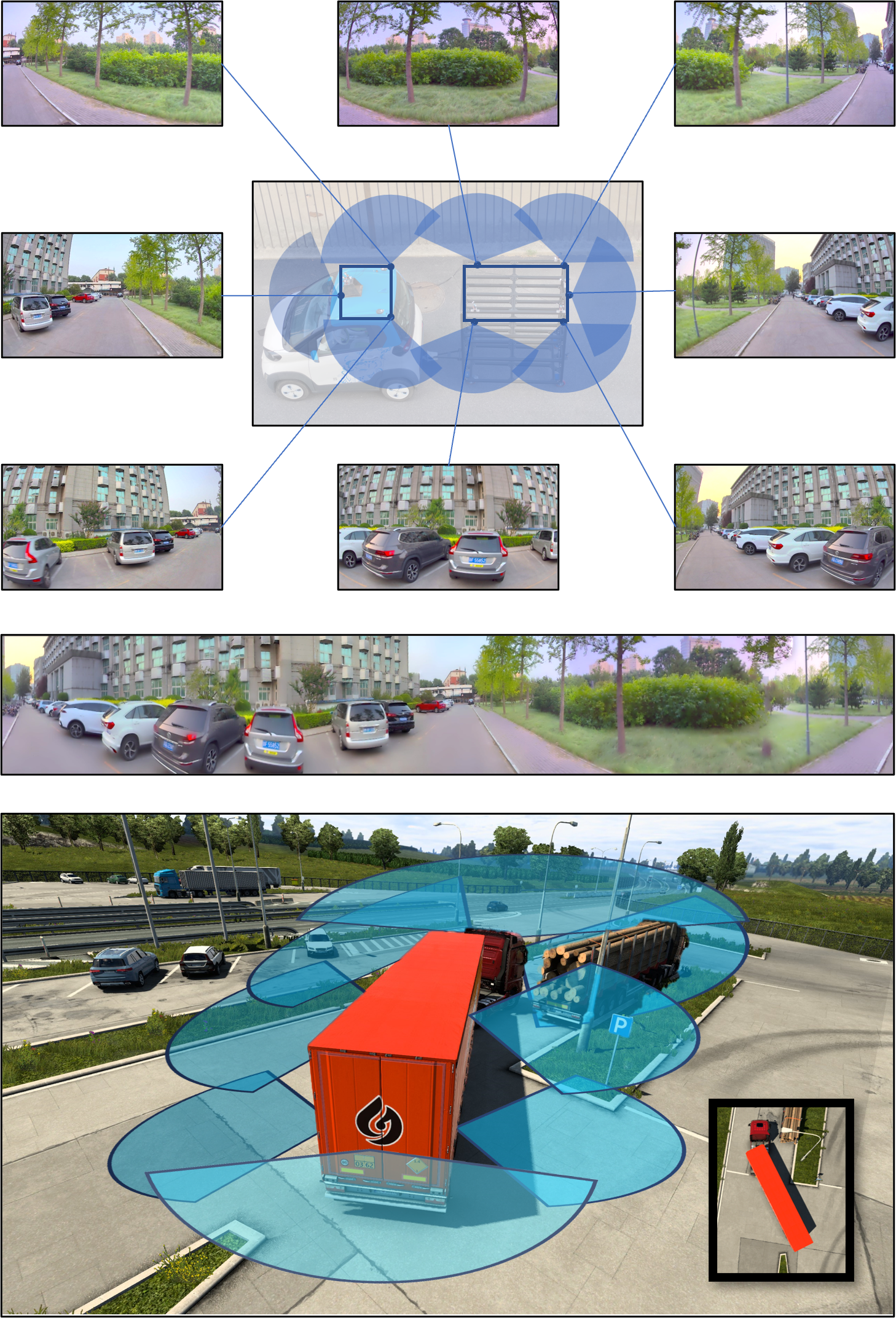}
\caption{Surround-view system of tractor-trailer wheeled robots.}
\label{fig:surroundview}
\end{figure}

% 常见的环视感知系统往往针对的是小型机器人或小型商用车辆，这意味着环视相机往往是固连在机械结构上的，相机之间的相对位置在初始布置时就已经确定，无法改变。这种固定的相机设置方案给机器人视觉感知带来了很大的好处，例如固定的相机配置只需要进行离线的一次性相机标定，固定的相机位姿可以为系统方程提供约束，紧凑的相机布置可以满足拼接同光心假设。然而，在面对大型拖车车辆的情况下，环视感知系统的摄像头必须均匀分布在tractor与trailer上，以解决巨大的车辆尺寸带来的盲区问题。因此，大型拖挂车的环视系统必然面临以下几个问题：首先，由于尺寸原因导致拖挂车的车头车身是由铰链连接的，这意味着tractor和trailer之间的相机相对位姿随时可能因为车辆转弯等情况发生剧烈变化。相机的相对布置不仅不能对系统方程提供预先的帮助，还成为了一个需要解决的未知量。其次，车头车身之间的非刚性连接导致其车头车身在行驶过程中的震动往往是不同的，这意味着车头车身上的相机震动的幅度与频率相互独立且会有较大的差异，这对相机的联合稳定与拼接的时间平滑性提出了较大的挑战。最后，由于拖挂车的尺寸往往是小型车辆的数倍，在相同相机配置下，其往往不能满足相机拼接的同一光心假设，大视差的引入会给拼接带来难度。
Common surround-view perception systems are often designed for small-scale robots or compact commercial vehicles, implying that the surround-view cameras are fixedly mounted on the mechanical structure, and the relative pose between cameras \textcolor{black}{is} predetermined during the initial setup, which cannot be changed. This fixed camera configuration offers significant benefits for robot visual perception. For instance, a fixed camera setup requires only one-time offline camera calibration\citep{calibration}, and the fixed camera poses provide constraints for the system equations, whereas the compact camera arrangement satisfies the parallax-free assumption for image stitching. However, when dealing with TTWRs, the surround-view system's cameras must be uniformly distributed on both the tractor and trailer parts to address blind-spot problems caused by the extensive  size\citep{paraidtrailer}\citep{control}. The common surround-view system of TTWRs is illustrated in Fig. \ref{fig:surroundview}.

% todo：分别根据三个挑战，分析问题并加国内外现状
Therefore, surround-view systems for TTWRs inevitably face several challenges. Firstly, the non-rigid connection between the tractor and trailer leads to different vibrations on the cameras of both parts, which are independent in amplitude and frequency\citep{traffic}\citep{airstabilization}. This introduces significant challenges for integrating image stabilization into the surround view system, as it must encompass both the tractor and the trailer. \textcolor{black}{The asynchronous vibrations between the tractor and trailer cause conflicting motion estimations for the same object in overlapping regions. Previous methods (e.g., UDIS2\citep{UDIS2}, Nie et al.\citep{stabstitch}) employ unidirectional motion estimation: designating one camera as a reference frame (mainly tractor view) and estimating relative displacements for the other. This imposes a static constraint on the reference view, forcing its motion trajectory to be modeled as zero. To address this, we design a Dual Independence Stabilization Motion Fields Estimation method to independently characterize motions for both units. By propagating these fields through joint optimization, we achieve bidirectional motion consistency while retaining maximum information fidelity.}

Secondly, due to the connection between the tractor and trailer \textcolor{black}{being} articulated, the relative camera poses between the tractor and trailer can drastically change during movements such as turns\citep{cameratrailer}\citep{vavm}. \textcolor{black}{Relative pose changes between the tractor and trailer during maneuvers (e.g., turning) cause dynamic misalignments in overlapping regions. Previous methods (e.g., UVSS\citep{UVSS})estimate relative tractor-trailer poses using mechanical sensors or visual odometry, then project images onto predefined geometric models (sphere/cylinder) based on relative pose estimation. This decoupled approach made pose estimation errors propagate to projection models, causing misalignment amplification. To address this, we propose a Random Plane based Stitching Motion Fields Estimation method that directly derives pixel-level displacements from feature motions without poses estimation and geometric priors. This method dynamically models spatial deformations as a stochastic adaptive plane rather than relying on rigid geometric priors. This innovation enables real-time compensation for arbitrary pose variations while preserving spatial coherence.}

Lastly, the considerable size difference between TTWRs and small robots makes it difficult to satisfy the parallax-free assumption for image stitching with the same camera configuration, resulting in large disparities that complicate the stitching process\citep{parallax2022}. \textcolor{black}{Previous methods (e.g., Guo et al.\citep{joint}) employ sequential pipelines that independently estimate stitching trajectories and stabilization trajectories and then fuse them through late-stage optimization. The independent estimation of stitching and stabilization trajectories inherently ignores their physical interdependence. To address this, we propose a Unified Vertex Motion Estimation method that estimates spatiotemporal motions, including those in non-overlapping regions. The corresponding joint optimization formulation considers the combined effects of stitching and stabilization.
}

% 为了解决以上问题，现有的方法大概分为两类。第一类方法将传感器集中在拖车车头上，这意味着其可以使用类似于传统机器人视觉环视系统方案，而避免拖挂车带来的上述问题。虽然这类方法针对拖车车距宽、转弯半径大等特点做了特殊的优化，但挂车传感器缺失使其难以完成拖挂车的环视感知，方法的重点在于拖挂车的规划控制。第二类方法使用非固连的移动相机，例如手持移动相机，同时进行拼接与防抖。但这类方法无法模拟车辆运行过程中的高频振动，也无法模拟车辆直行与转弯给相机位姿带来的随机分布。同时，这些方法都将拼接模块和防抖模块看为独立的管线，在拼接和防抖计算结束后，通过联合优化的方法进行联合。这样的流程使得拼接和防抖的结合不够紧密。
% To address these challenges, existing methods can be broadly classified into two categories. The first category of methods concentrates the sensors on the tractor, allowing them to adopt traditional robot vision-based surround-view system schemes while avoiding the aforementioned problems arising from the trailer\citep{traffic}\citep{stabil}. Although these methods are optimized for the unique characteristics of TTWRs, such as large size and large turning radius, the absence of sensors on the trailer makes it difficult to achieve surround-view perception for the entire robot. The focus of these methods lies in the planning and control of TTWRs. The second category of methods employs non-rigidly connected moving cameras, such as handheld cameras, for both image stitching and stabilization\citep{joint}. However, these methods are not designed for the high-frequency vibrations experienced by cameras and the random distribution of camera poses caused by non-rigid connections. Additionally, these methods treat the stitching and stabilization modules as separate pipelines and perform joint optimization only after completing both tasks. This workflow leads to less cohesive integration between image stitching and stabilization.

To achieve a comprehensive solution, a novel approach is needed that can effectively handle the challenges posed by TTWRs and ensure tight integration between image stitching and stabilization.

% 在这篇文章中，我们认为解决抖动问题的关键是估计intra motions (motions within a video between neighboring frames)；类似的，解决拼接问题的关键是估计inter motions (motions at the corresponding frames between different videos)[guo]。对于抖动问题来说，估计intra motions可以通过平滑消除其因抖动导致的视角跳变，从而实现稳定效果；对于拼接问题来说，估计inter motions可以通过最小化像素重投影误差消除重叠区域的伪影，从而实现拼接。估计这两种运动使得在同一框架下同时实现稳定与拼接成为可能，具体来说，防抖的目的是最小化前后帧相同特征的距离，这是在时间维度下对图像变化的约束；防抖的目的是最小化左右帧相同特征的距离，这是在空间维度下对图像变化的约束。在统一的时空间维度下，在完成空间相邻帧、时间相邻帧数据关联后即对图像进行约束，即可实现拼接防抖的紧密结合。另外，我们引入了一种新的联合优化公式，考虑了拼接和稳定的综合效果，以此来应对size导致的camera parallax
We consider that the key to solving the problem of asynchronous vibrations lies in estimating intra motions (motions within a video between neighboring frpames), the key to solving the problem of stitching being unable to adapt to pose changes lies in estimating inter motions (motions at corresponding frames between different videos)\citep{joint}. Estimating intra motions can effectively smooth out abrupt changes in viewpoints caused by vibration, thereby achieving stabilization. On the other hand, estimating inter motions can minimize pixel reprojection errors in the overlapping regions, thus achieving seamless image stitching. Estimating these two types of motion allows us to achieve stabilization and stitching within the same framework. Specifically, the goal of stabilization is to minimize the distance between matched features in consecutive frames, which imposes temporal constraints on image variations. Meanwhile, the goal of stitching is to minimize the distance between matched features in adjacent frames, which imposes spatial constraints on image variations. By unifying these temporal and spatial dimensions, we can seamlessly integrate stabilization and stitching by constraining the images based on data associations in both spatially adjacent and temporally adjacent frames. Additionally, we have introduced a novel joint optimization function that considers the combined effects of stitching and stabilization to address camera parallax caused by large size.

In this paper, we propose a unified vertex motion video stabilization and stitching framework for the surround-view systems of tractor-trailer wheeled robots. To strike a balance between alignment accuracy and computational efficiency, we adopt a strategy similar to MeshFlow \citep{meshflow}, which exclusively estimates the displacement of mesh vertices in image coordinates. For intra-frames (different temporal frames within a video), the displacement of each vertex will be temporally accumulated to estimate camera motion and then smoothed to stabilize the videos. Concerning inter-frames between adjacent cameras, the detected feature motions will be propagated to the vertices of corresponding frames, and the estimated motion field will guide the alignment of matching frames. Temporal trajectories and stitching profiles will be jointly optimized to generate a steady \textcolor{black}{surround-view} panorama.

% Following our preliminary exposition of core concepts at the 2023 IEEE/RSJ International Conference on Intelligent Robots and Systems (IROS) \cite{UVSS}, this manuscript substantially expands upon our prior work. On one hand, this submission introduces an innovative mesh-based unified vertex motion model that concurrently addresses the challenges of video stabilization and stitching. On the other hand, our experimental investigations utilize data directly from real-world tractor-trailer vehicles, marking a significant departure from the small-scale model vehicles employed in earlier studies.

In brief, our main contributions are as follows:
\begin{itemize}
% 我们提出了一种新的mesh-based motion model，可以同时表征防抖的motion与拼接的motion。从而在时空间维度下统一防抖和拼接带来的图像变化。
% 我们提出了一种新的联合优化方程，考虑到拼接与防抖对优化方程的联合影响，我们将总代价近似为拼接代价与防抖代价的加权和，这意味着可以通过调节权重来增强拼接或增强防抖效果。
% 我们完成了一套可以实际在拖挂车上运行的环视系统软硬件方案，并进行了实验。据我们所知，这是第一套针对拖挂车环视图像拼接与防抖的技术方案。
\item \textcolor{black}{We design a Dual Independence Stabilization Motion Fields Estimation method that independently characterizes motion patterns of the tractor and trailer units. Through joint optimization propagation, it achieves bidirectional motion consistency while preserving maximum information fidelity, effectively addressing object ambiguities induced by asynchronous vibration patterns.}
\item \textcolor{black}{We propose a Random Plane based Stitching Motion Estimation method, which directly derives pixel-level displacements from feature motions without relying on pose estimation or geometric priors. By modeling spatial deformations as stochastic adaptive planes, this method resolves dynamic ghosting artifacts caused by articulated pose variations.}
\item \textcolor{black}{We develop a Unified Vertex Motion Estimation method, which estimates spatiotemporal motions, including those in non-overlapping regions. The corresponding joint optimization framework accounting for combined stitching-stabilization effects solves the problem of distortion propagation through low-overlapping regions that cause catastrophic geometric distortion amplification.}
\item We have implemented a practical hardware and software solution for a surround-view system that can be deployed on tractor-trailer wheeled robots. Extensive experiments were conducted to validate the proposed approach.
\end{itemize}

\section{Related Work}
\label{related}
In this section, we provide a comprehensive overview of the existing literature on image stitching, image stabilization, and their integration.
\subsection{Image stitching}
Image stitching is a computational technique aimed at seamlessly merging multiple partially overlapping scene images to create a single image encompassing a wider field of view and enhanced resolution. In the early stages of image stitching research, global homography-based methods were commonly employed \citep{automatic}. However, contemporary approaches have evolved to incorporate a combination of local homography estimation \citep{SPHP}, \citep{AANAP}, and seam-driven optimization techniques \citep{seam}, yielding superior results.

J. Zaragoza et al. \citep{APAP} introduce a novel mesh-based image deformation and alignment strategy known as "as-projective-as-possible" (APAP). This technique effectively aligns overlapping regions and preserves the shape of stitched images through advanced estimation methods. On the other hand, the approach presented in \citep{parallax} focuses on optimizing a seam using minimal energy to address local misalignment caused by significant parallax. Furthermore, leveraging state-of-the-art deep learning techniques, Nie et al. \citep{nie_unsupervised} propose a comprehensive framework for image stitching based on unsupervised learning. This framework includes coarse image alignment and unsupervised image reconstruction, pushing the boundaries of image stitching algorithms.
Fu et al. \citep{lianghao} divide the image stitching methods into two categories, namely, mosaic stitching methods for generating stitched plane images and panoramic stitching methods for generating stitched panoramic images. 

These approaches have made notable advancements in the field of image stitching. However, they face challenges when the camera poses undergo significant variations, such as when a truck makes sharp turns, leading to substantial changes in the overlapping regions between the truck's front and body. Consequently, effectively addressing the problem of drastic changes in overlap caused by varying angles remains an unresolved challenge.
% 为了解决以上问题，我们将图像拼接视为空间模块，通过实时计算图像位姿，解算图像空间运动轨迹，并将其送入后面结构进行统一处理。

\subsection{Image stabilization}
The existing techniques for video stabilization can be broadly categorized into two groups: traditional methods and learning-based methods. Traditional methods primarily aim at recovering and subsequently smoothing the camera's motion to achieve video stabilization. In contrast, learning-based methods leverage the learning capabilities of neural networks to achieve end-to-end generation of stable frames.

Traditional methods in video stabilization can be categorized into 2D, 3D, and 2.5D approaches based on the dimension of the estimated camera trajectories. 2D methods typically involve calculating linear transformations between adjacent frames \citep{2006full}, followed by the \textcolor{black}{application} of low-pass filters \citep{2006robust} or L1-norm optimizers \citep{L1} to achieve parameter smoothing. On the other hand, 3D methods \citep{3Dcontent} estimate 3D camera paths using structure from motion (SfM) techniques. Subspace video stabilization \citep{subspace} decomposes the scene and camera motion using subspace constraints and achieves trajectory smoothing through polynomial fitting. While 3D methods generally yield superior stabilization results, they may suffer from limited robustness and efficiency. In contrast, 2.5D methods, exemplified by bundled paths stabilization \citep{bundled}, divide images into uniform meshes and independently calculate 2D trajectories. Liu et al. propose MeshFlow \citep{meshflow}, which propagates feature motions to mesh vertices, effectively preserving the spatial coherence of stabilized frames. 

Learning-based methods in video stabilization have emerged as a promising direction. StabNet \citep{stabnet} is a pioneering approach that utilizes convolutional neural networks for video stabilization. Its training process leverages historically stabilized frames and supervised ground-truth steady frames. DIFRINT \citep{difrint}, proposed by Choi et al., employs video interpolation techniques to stabilize videos whereas mitigating ghosting artifacts from moving objects and distortion caused by parallax. Xu et al. introduce DUT \citep{dut}, which integrates deep learning into the pipeline of traditional methods by enhancing sub-modules. By leveraging self-supervised learning on a collection of unsteady videos, learning-based methods achieve favorable performance in stabilizing frames\citep{nndvs}.

These approaches have made notable advancements in the field of image stabilization. However, these methods only require intra-motion information for estimating camera paths, as opposed to relying on long feature trajectories. 

\subsection{Video Stitching and Stabilization}
For multiple videos captured by dynamic cameras, previous works such as \citep{panoramic} and \citep{hamza2015} have focused on stabilizing panoramic videos that have already been stitched. In contrast, Lin et al. \citep{lin2016seamless} address the stabilization of individual steady videos and incorporate dense 3D reconstruction techniques. However, this approach introduces significant computational complexity and may render the stitching process more fragile.

To tackle these challenges, Guo et al. \citep{joint} propose a method that enforces stitching as a rigid constraint during the optimization of stabilization variables. This approach generates a common smooth camera path for all input videos, ensuring consistent stitching and stabilization results. Similarly, Nie et al. \citep{dynamic} perform a joint optimization of both stabilization and stitching variables, aiming to make the new camera paths as similar as possible to their original duplication. Furthermore, Nie et al. \citep{stabstitch} have simultaneously achieved video stitching and stabilization within a unified unsupervised learning framework.

These approaches have made notable advancements in the field of video stitching and stabilization. However, these methods typically treat the stitching module and the stabilization module as separate pipelines, where the stitching and stabilization computations are performed independently and then combined through joint optimization, resulting in a loose integration of stitching and stabilization.

In our work, we propose a different perspective, considering stitching and stabilization as two variations of the same underlying task. We recognize that stitching involves aligning spatial data from adjacent frames captured at the same time, while stabilization involves aligning temporal data from consecutive frames captured by the same camera. By unifying these two tasks within a single processing pipeline, we can simultaneously achieve both stitching and stabilization, taking advantage of their shared characteristics and improving the overall integration of the two processes.

\begin{figure}[!t]
\centering
\includegraphics[width=\linewidth]{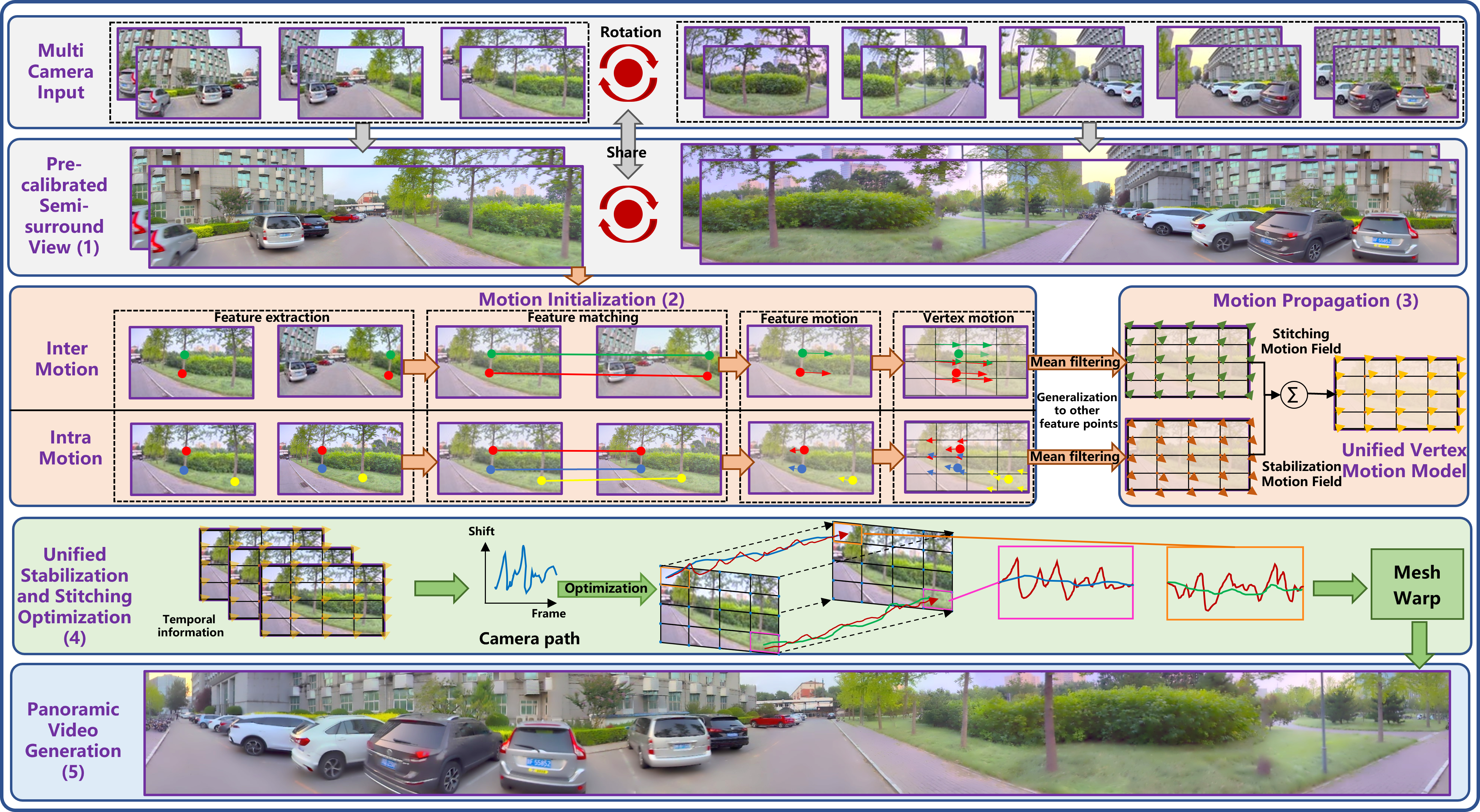}
%Overview of our method.数据来自我们在北京理工大学采集的实景。方法流程大致分五部分：（1）Pre-calibirated semi-surround view:将固连在trator的相机图像和trailer的相机图像分别进行拼接，采用预先标定好的参数，形成前视的广角图像与后视的广角图像；（2）motion initialization:通过特征匹配计算特征点之间的位移并推广至相应的网格顶点上，相同时刻相邻相机是inter motion ，相邻时刻同一相机是intar motion；(3)motion Propagation:对同一顶点上由不同特征点推至的motion进行滤波操作，使得每个顶点只有一个motion，然后叠加形成unified vertex motion；（4）unified stabilizaiton and stitching optimizaiton：对motion叠加形成的轨迹进行平滑，总代价近似为拼接代价与防抖代价的加权和，这意味着可以通过调节权重来增强拼接或增强防抖效果；（5）panoramic video generations：将优化后的图像进行整合，生成全景图。
\caption{\textbf{Overview of our method:} Leveraging real-scene data from Beijing Institute of Technology, our method unfolds across five stages: (1) Pre-calibrated Semi-Surround View: Merging images from tractor and trailer cameras using \textcolor{black}{pre-calibrated} parameters to form forward and rearward wide-angle views; (2) Motion Initialization: Computing feature point displacements, distinguishing inter-motion and intra-motion for adjacent and same cameras at varying moments; (3) Motion Propagation: Filtering and amalgamating motions on vertices to craft a unified vertex motion; (4) Unified Stabilization and Stitching Optimization: Smoothing motion trajectories, tuning weights to balance stitching and stabilization costs; (5) Panoramic Video Generation: Integrating optimized images to craft a panoramic view.}
\label{fig:system}
\end{figure}

\section{Our Approach}
\label{approach}

\begin{figure}[!t]
\centering
\includegraphics[width=\linewidth]{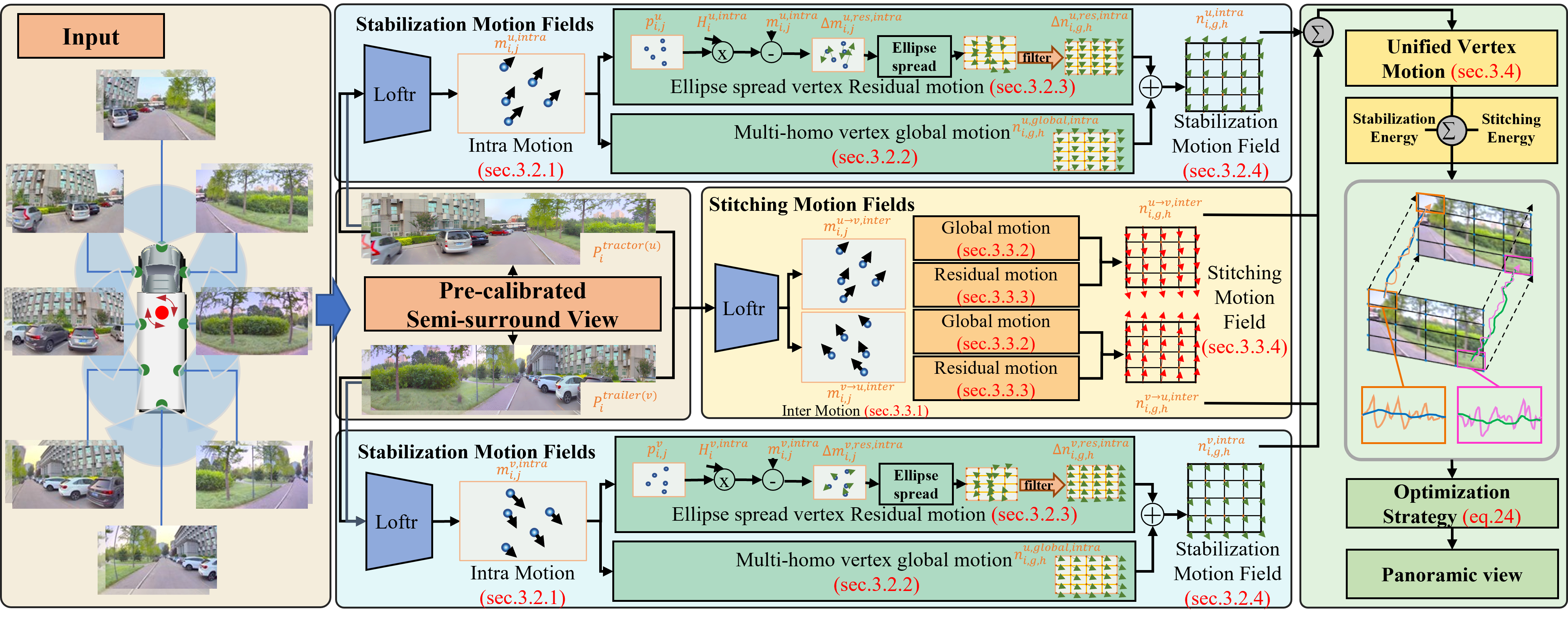}
\caption{\textcolor{black}{\textbf{Overview of Specific Technical Aspects:} In the diagram, orange symbols represent data, $\textcolor{orange}{m_{i,j}^{u,\text{intra}}}$ (intra-unit feature motion), 
$\textcolor{orange}{m_{i,j}^{u \to v}}$ (inter-unit correspondence motion). 
\textcolor{black}{Black annotations} describe core modules: 
\textit{Stabilization Motion Fields} for temporal vibration compensation, 
\textit{Stitching Motion Fields}  for spatial alignment.}}
\label{fig:pipeline}
\end{figure}

A schematic overview of the unified vertex motion Video Stabilization and Stitching method is presented in Fig.\ref{fig:system}. Our approach is primarily based on a unified optimization framework that simultaneously addresses video stitching and stabilization tasks.  In the initial phase, camera units installed on both the tractor and the trailer undergo separate pre-calibration and individual stitching procedures to produce two partial surround views of the robot. Specifically, one partial surround view is constructed from the tractor-mounted camera, while the other is generated from the trailer-mounted camera. Following this, feature matching is employed to establish intra motions and inter motions, thus forming a unified vertex motion, which serves to harmonize the motion associated with both stabilization and stitching. %Then, the unified optimization process is aimed at mitigating temporal instability within the intra-frames of each camera and aligning the inter-frames between the two adjacent partial surround views.
\textcolor{black}{Then, the unified optimization process simultaneously addresses temporal stabilization (smoothing intra-camera motions across consecutive frames) and spatial stitching (aligning inter-camera views from adjacent tractor/trailer units), thereby enforcing spatiotemporal consistency in both overlapping and non-overlapping regions.}
Lastly, the panoramic video is artfully crafted through mesh warping methods, akin to those delineated in \citep{single}. A detailed block diagram showcasing the specific technical aspects is presented in Fig. \ref{fig:pipeline}.
\textcolor{black}{\subsection{Pre-calibrated Semi-surround View}
The tractor-trailer system deploys a distributed camera array $\mathcal{C} = \{c^k \, | \, k \in [1, 8]\}$ with 3 tractor-mounted ($c^{1:3}$) and 5 trailer-mounted ($c^{4:8}$) cameras. Each camera stream is denoted as $\mathcal{F}^k = \{f_i^k \in \mathbb{R}^{H \times W \times 3} \, | \, i \in [1, N]\}$, where $f_i^k$ represents the $i$-th frame from camera $k$.
For each articulated unit (tractor/trailer), we compute static homographies:
\begin{equation}
\forall u \in \{\text{tractor}, \text{trailer}\}, \quad \mathbf{H}^u = \arg\min_{\mathbf{H}} \sum_{k,m \in u} \| \mathbf{K}_k^{-1}\mathbf{p}_k - \mathbf{H}_{m \to k} \mathbf{K}_m^{-1}\mathbf{p}_m \|^2
\end{equation}
where $\mathbf{p}_k \in \mathbb{R}^2$ denotes the image coordinates of checkerboard corners or features in camera $k$, $\mathbf{H}_{m \to k}$ aligns adjacent cameras within unit $u$, and $\mathbf{K}_k$ are pre-calibrated intrinsics. This generates the $i$-th partial panoramas:
\begin{equation}
P_i^{\text{tractor}} = \bigoplus_{k=1}^3 \mathcal{W}(f_i^k; \mathbf{H}^{\text{tractor}}), \quad P_i^{\text{trailer}} = \bigoplus_{k=4}^8 \mathcal{W}(f_i^k; \mathbf{H}^{\text{trailer}})
\end{equation}
where $\mathcal{W}(\cdot)$ denotes homography warping and $\bigoplus$ is multiband blending.
\subsection{Stabilization Motion Fields Estimation}
\label{sec:stab_fields}
To address asynchronous vibrations between tractor and trailer units, we propose a Dual Independence Stabilization Motion Fields Estimation method to independently characterize motions for both units.
\subsubsection{Intra Motion Initialization} \label{sec:intra_motion}
For each partial view $P_i^u$ (where $u \in \{\text{tractor}, \text{trailer}\}$), we extract sparse feature matches between consecutive frames using LoFTR\citep{loftr}:
\begin{equation}
\{ \mathbf{p}_{i,j}^u, \mathbf{p}_{i+1,j}^u \} = \text{LoFTR}(P_i^u, P_{i+1}^u), \quad j \in [1, M]
\end{equation}
}

Here, \textcolor{black}{$\lbrace {p_{i,j}^{u},{{p}}_{i+1,j}^{u}} \rbrace$ }represents the $j$-th matched feature pairs in intra frames. \textcolor{black}{${p_{i,j}^{u}}$} represents the coordinates of the \(j\)-th matched feature in the \(i\)-th frame from \textcolor{black}{the \(u\) view.}
For intra-frames, we directly utilize two adjacent frames as input. 
%However, for inter-frames, only the consensus region among cameras is processed to enhance computational efficiency and prevent potential mismatches caused by duplicate textures.

%LoFTR is a novel method for local feature matching between image pairs, which has been widely used in some other vision tasks [], []. 
Compared to traditional feature matching pipelines, LoFTR differentiates itself by not responding on artificial descriptors and instead leveraging image pairs as input. Using the global vision provided by Transformers and employing a coarse-to-fine refinement module, LoFTR is capable of generating dense matches even in low-texture areas, where traditional feature detectors or other deep learning methods \citep{superpoint} tend to struggle. 

\textcolor{black}{Subsequently, the intra motion vector is computed via robust RANSAC filtering:
\begin{equation}
\mathbf{m}_{i,j}^{u,intra} = \text{MAGSAC}(\mathbf{p}_{i+1,j}^u - \mathbf{p}_{i,j}^u)
\end{equation}
}
where \textcolor{black}{${m}_{i,j}^{u,intra}$ is the intra motion vector of the $j$th matched feature in the \(i\)-th frame from the \(u\) view.}

\begin{figure}[!t]
\centering
\subfloat[Intra motion]{\includegraphics[width=0.32\linewidth]{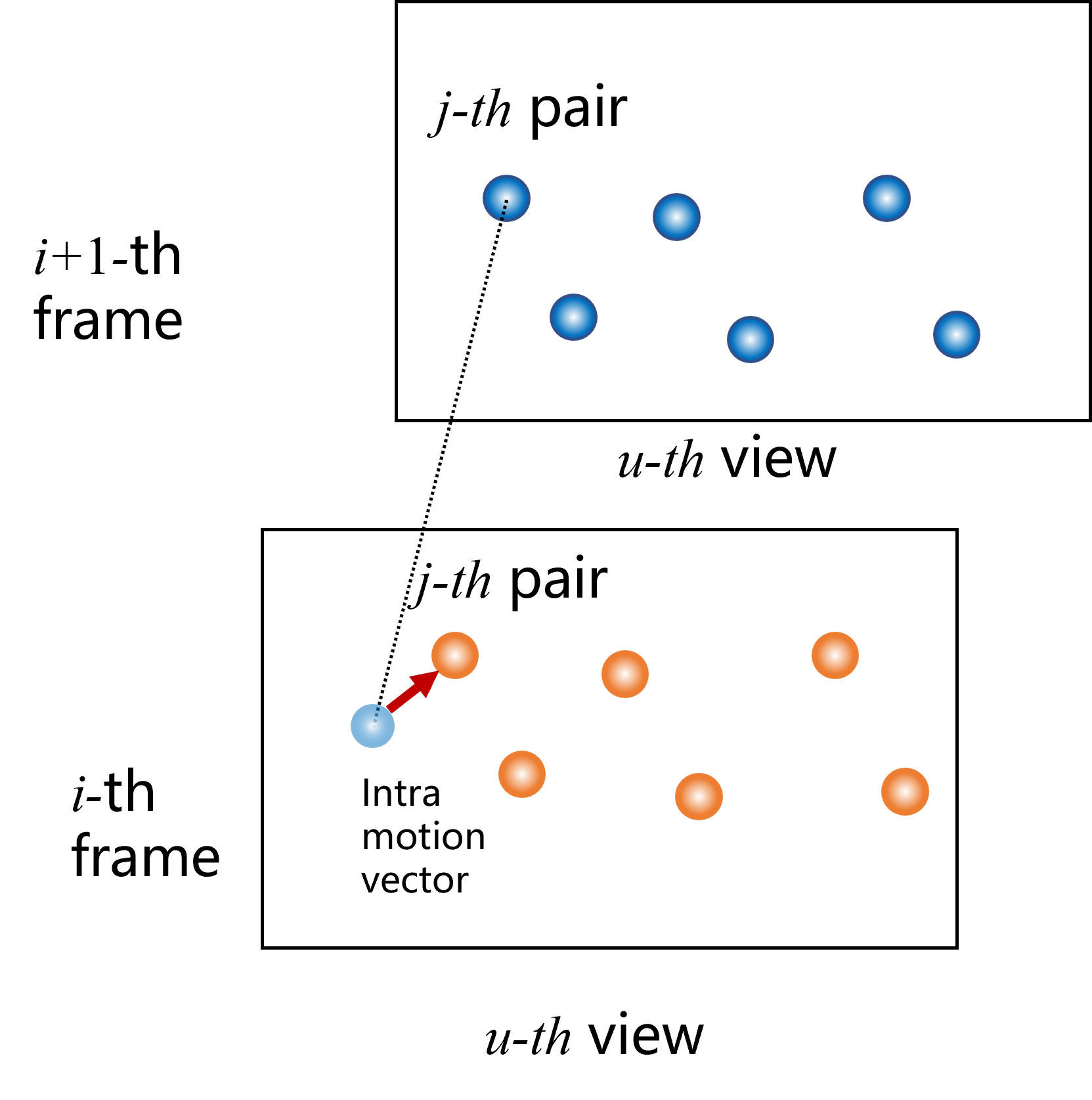}}
\hfil
\subfloat[Inter motion]{\includegraphics[width=0.5\linewidth]{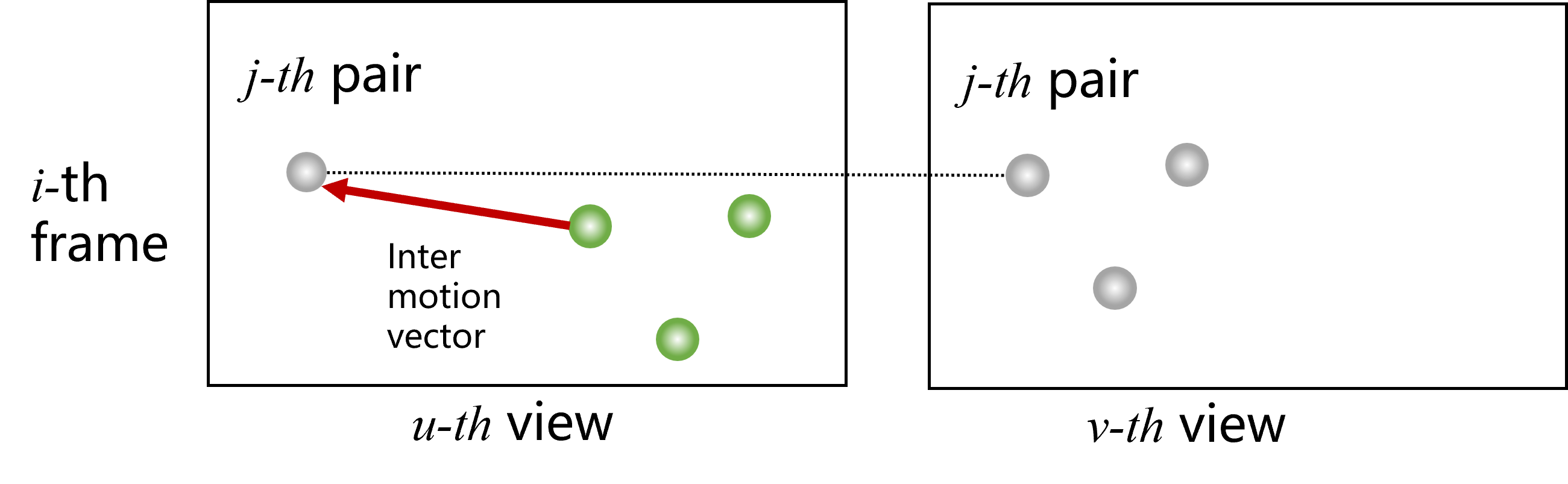}}
\caption{\textcolor{black}{\textbf{Illustration of Intra Motion and Inter Motion:} Demonstrating the distinction between Inter motion and Intra motion, with Inter motion showcasing the movement between adjacent views at a singular moment and Intra motion depicting the movement within the same view across adjacent moments.}}
\label{fig:featuremotion}
\end{figure}

We employ MAGSAC++ \citep{magsac++}, a robust estimator, to mitigate the undesirable effects caused by mismatches and dynamic objects. The iterative reweighted least squares optimization of MAGSAC++ exhibits excellent performance with a soft threshold, making it highly practical and suitable for a wide range of scenarios. The schematic representation of the Intra motion is depicted in Fig. \ref{fig:featuremotion}.

\textcolor{black}{\subsubsection{Global Intra Motion}
\label{glbal sta}
To decouple the dominant camera motion from local vibrations, we decompose intra motions into \textit{global motion} ( estimated by homography matrices that minimize reprojection errors across all features) and \textit{residual motion} (capturing local deformations unexplained by the global motion). This hierarchical decomposition enables robust stabilization while preserving the structure of the scene.}

\textcolor{black}{\textbf{Feature-Level Global Intra Motion:}
For each matched feature pair $\{ \mathbf{p}_{i,j}^u, \mathbf{p}_{i+1,j}^u \}$ in unit $u$, we estimate:
\begin{equation}
\label{equ:sta glo}
\underbrace{\mathbf{m}_{i,j}^{u,\text{global},intra}}_{\text{Global Intra Motion}} = \mathbf{H}_i^{u,intra}(\mathbf{p}_{i,j}^u) - \mathbf{p}_{i,j}^u, \quad \underbrace{\Delta\mathbf{m}_{i,j}^{u,\text{res},intra}}_{\text{Residual Intra Motion}} = \mathbf{m}_{i,j}^{u,\text{intra}} - \mathbf{m}_{i,j}^{u,\text{global},intra}
\end{equation}
where $\mathbf{H}_i^{u,intra} \in \mathbb{R}^{3\times3}$ is the global homography matrix for unit $u$ at frame $i$, computed via RANSAC:
\begin{equation}
\mathbf{H}_i^{u,intra} = \arg\min_{\mathbf{H}} \sum_{j=1}^M \rho\left( \| \mathbf{H}(\mathbf{p}_{i,j}^u) - \mathbf{p}_{i+1,j}^u \|^2 \right)
\end{equation}
with $\rho(\cdot)$ being the MAGSAC++ robust cost function.
}

\textcolor{black}{\textbf{Vertex-Level Global Intra Motion:}}
After the initialization of intra motions, \textcolor{black}{each partial view $P_i^u$ is partitioned into an $m \times n$ grid of meshes $\{ \Omega^g \, | \, g \in [1, mn] \}$. Each mesh $\Omega^g$ contains four vertices $\{ v_{g,h} \in \mathbb{R}^2 \, | \, h \in \{1,2,3,4\} \}$ at its corners.} This approach strikes a balance between computational efficiency and spatial smoothness, as it avoids the computational overhead associated with dense methods whereas improving the overall visual coherence of the results.

% todo:解释vertex推广与global motion 和相对motion
% 类似于【meshflow】，如图\ref Fig4所示，我们将特征点上的motion，推广到椭圆内的所有顶点上。在motion Propagation以前，我们给顶点分配一个global motion，其由传统homography计算得来。显然顶点motion为global motion与residual motion的叠加。相似的思想也在cite[zhang2023minimum]中用于视频稳定，在文献cite[单应性]中用于单应性估计。对于一些没有特征被检测到的极端情况（地面、天空），分配全局运动给网格顶点，防止motion的缺失。在此基础上，我们将the motion at the vertices分为the sum of global motion and residual motion
%Drawing inspiration from Meshflow, we extend the motion present on the feature points to all vertices within the ellipse, as depicted in Fig. \ref{fig:vertmotion}.

\textcolor{black}{Building on the foundation of decomposition, we also decompose the motion at the vertices into the sum of global motion and residual motion.} The global motion is computed using traditional homography. For extreme cases where no features are detected (such as on the ground or sky), global motion is allocated to the mesh vertices to prevent the absence of motion.

\begin{figure}[!t]
\centering
\includegraphics[width=\linewidth]{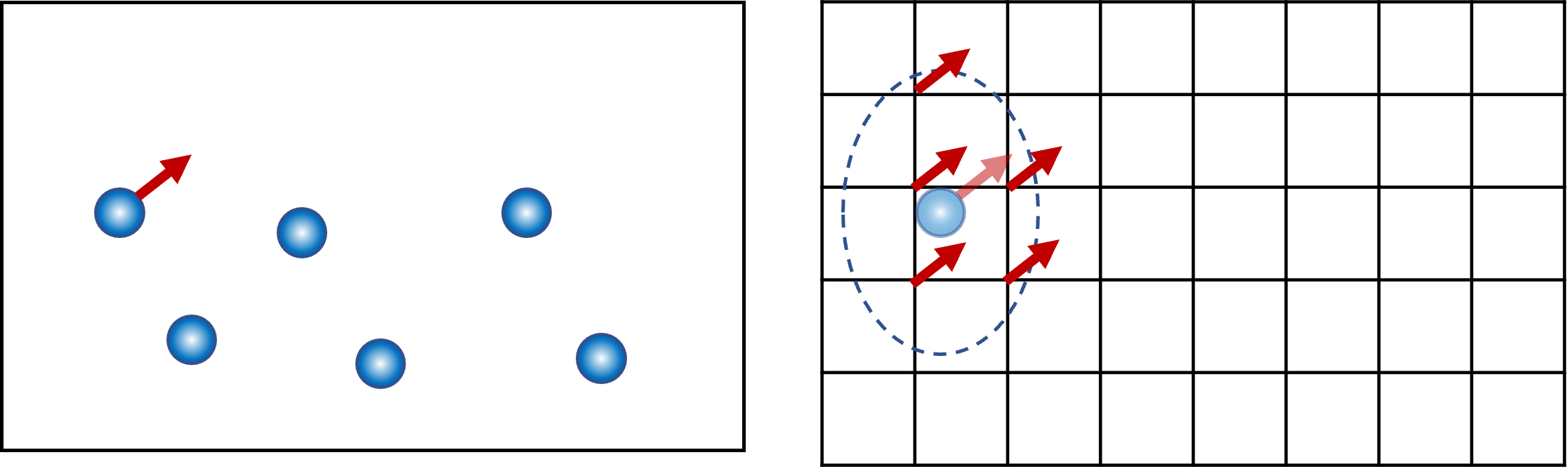}
\caption{\textbf{Illustration of vertex motion:} This illustration delineates the concept of vertex motion, detailing the extrapolation of motion from feature points to all vertices within the ellipse.}
\label{fig:vertmotion}
\end{figure}

\textcolor{black}{For each mesh $\Omega^g$, we compute a local homography $\mathbf{H}_{i,g}^u$ using intra-feature motions within the mesh:
\begin{equation}
\mathbf{H}_{i,g}^{u,intra} = \arg\min_{\mathbf{H}} \sum_{\mathbf{m}_{i,j}^{u,\text{intra}} \in \Omega^g} \rho\left( \| \mathbf{H}(\mathbf{p}_{i,j}^u) - \mathbf{p}_{i+1,j}^u \|^2 \right)
\label{eq:mesh_homo}
\end{equation}
where $\rho(\cdot)$ is the MAGSAC++ robust cost function.}

\textcolor{black}{The global intra motion of each vertex $v_{g,h}$ is derived by applying the mesh homography:
\begin{equation}
\mathbf{n}_{i,g,h}^{u,\text{global},intra} = \mathbf{H}_{i,g}^{u,intra}(v_{g,h}) - v_{g,h}
\label{eq:vertex_global}
\end{equation}
}

%
% For inter-frames, the presence of moving objects is considered to have limited effects on stitching. Therefore, the global vertex motion can be simplified as:

% \begin{equation}
% \setlength{\abovedisplayskip}{-3pt}
%     \hat{n}_{i,g}^{inter} = H_{i}( v_{i,g}) - v_{i,g}
% \end{equation}

% In this equation, $H_{i}$ represents the global homography obtained from traditional homography matrix estimation methods (after obtaining the feature point pairs between two images, the global homography matrix can be easily derived).$\hat{n}_{i,g}^{inter}$ represents the global vertex motion of the $g$th vertex within the mesh in $i$th the frame, specifically related to inter-frame motion.

\textcolor{black}{\subsubsection{Residual Intra Motion Compensation}
\textbf{Feature-Level Residual Intra Motion:}For each matched feature pair $\{ \mathbf{p}_{i,j}^u, \mathbf{p}_{i+1,j}^u \}$ in unit $u$, the residual motion ${\Delta\mathbf{m}_{i,j}^{u,\text{res},intra}}$ is calculated in equ.\ref{equ:sta glo}.}

\textcolor{black}{\textbf{Vertex-Level Residual Intra Motion:}The residual vertex motion captures local deformations unexplained by the global homography:
\begin{equation}
\Delta\mathbf{n}_{i,g,h}^{u,\text{res},intra} = \sum_{j \in \mathcal{N}(g,h)} w_{j} \Delta\mathbf{m}_{i,j}^{u,\text{res},intra}
\label{eq:vertex_res}
\end{equation}
where $\mathcal{N}(g,h)$ denotes features in the ellipse of vertex $v_{g,h}$, and $w_j$ are Gaussian weights based on spatial distances.}

% 在生成了vertex motion的过程中，可以看到同一个顶点会受到其附近不同特征点带来的motion影响（正如上面所说，没有受到特征点影响的顶点会被分配global motion），但显然一个顶点同一时间只能有一个确定的motion。为了从多个特征点传播过来motion生成较为准确的vetex motion，additional modules are still necessary to accurately characterize the motion of each camera and the image transformation from non-fixed cameras.

During the process of generating vertex motion, it is observable that a single vertex can be influenced by the motions emanating from different nearby feature points. (As mentioned earlier, vertices not influenced by feature points are assigned a global motion.) Drawing inspiration from Meshflow, we extend the residual motion present on the feature points to all vertices within the ellipse, as depicted in Fig. \ref{fig:vertmotion}.

\subsubsection{Stabilization Motion Field}
\label{sec:stabmotion}
Videos with high frame rates often exhibit minimal parallax between intra-frames, estimating the motion field primarily focused on accuracy and mitigating the effects of moving objects. \textcolor{black}{By considering the global motion and residual motion of the vertices, the final motion of the vertices can be calculated as follows:
\begin{equation}
\mathbf{n}_{i,g,h}^{u,intra} = \underbrace{\mathbf{n}_{i,g,h}^{u,\text{global},intra}}_{\text{Global}} + \underbrace{\Delta\mathbf{n}_{i,g,h}^{u,\text{res},intra}}_{\text{Residual}}
\label{eq:final_motion}
\end{equation}
}

where ${n}_{i,g,h}^{intra}$ represents the stabilization motion of the $h$-th vertex within the $g$ mesh in $i$-th the frame, which is calculated in  (\ref{eq:final_motion}). The stabilization motion field will be accumulated frame by frame to simulate the camera path and smoothed in the next stage.

\textcolor{black}{
\subsection{Stitching Motion Fields Estimation} \label{sec:stitch_fields}
To achieve alignment between tractor and trailer units, we propose a Random Plane based Stitching Motion Fields Estimation method that directly derives pixel-level displacements from feature motions without poses estimation and geometric priors.
}
\textcolor{black}{
\subsubsection{Inter Motion Initialization} \label{sec:inter_motion}
For adjacent units $u$ (tractor) and $v$ (trailer), we extract cross-unit feature correspondences between their partial views using LoFTR:
\begin{equation}
\{ \mathbf{p}_{i,j}^u, \mathbf{p}_{i,j}^v \} = \text{LoFTR}(P_i^u, P_i^v), \quad j \in [1, K]
\end{equation}
where $\mathbf{p}_{i,j}^u \in P_i^u$ and $\mathbf{p}_{i,j}^v \in P_i^v$ are matched features in the overlapping region (Fig.~\ref{fig:featuremotion}b). Inter motion vectors are computed as:
\begin{equation}
\mathbf{m}_{i,j}^{u \to v,inter} = \text{MAGSAC}(\mathbf{p}_{i,j}^v - \mathbf{p}_{i,j}^u)
\end{equation}
The schematic representation of Inter motion is depicted in Fig. \ref{fig:featuremotion}.
}
\textcolor{black}{
\subsubsection{Global Inter Motion} \label{sec:stitch_global}
We also decompose inter motions into \textit{global motion} and \textit{residual motion}, as mentioned in sec \ref{glbal sta}.
}
\textcolor{black}{
\textbf{Feature-Level Global Inter Motion:}For each matched feature pair $\{ \mathbf{p}_{i,j}^u, \mathbf{p}_{i,j}^v \}$ in unit $u$, we estimate:
\begin{equation}
\label{equ:decom}
\underbrace{\mathbf{m}_{i,j}^{u \to v,\text{global},inter}}_{\text{Global Inter Motion}} = \mathbf{H}_i^{u \to v}(\mathbf{p}_{i,j}^u) - \mathbf{p}_{i,j}^u, \quad \underbrace{\Delta\mathbf{m}_{i,j}^{u \to v,\text{res},inter}}_{\text{Residual Inter Motion}} = \mathbf{m}_{i,j}^{u \to v,inter} - \mathbf{m}_{i,j}^{u \to v,\text{global},inter}
\end{equation}
The dominant alignment between units is modeled by a stitching homography $\mathbf{H}_i^{u \to v}$:
\begin{equation}
\mathbf{H}_i^{u \to v} = \arg\min_{\mathbf{H}} \sum_{j=1}^K \rho\left( \| \mathbf{H}(\mathbf{p}_{i,j}^u) - \mathbf{p}_{i,j}^v \|^2 \right)
\end{equation}
}
\textcolor{black}{
\textbf{Vertex-Level Global Inter Motion:}
The global inter motion of each vertex $v_{g,h}$ is then propagated:
\begin{equation}
\mathbf{n}_{i,g,h}^{{u \to v},\text{global},inter} = \mathbf{H}_i^{u \to v}(v_{g,h}^u) - v_{g,h}^u
\end{equation}
\subsubsection{Residual Inter Motion Compensation} \label{sec:stitch_residual}
\textbf{Feature-Level Residual Inter Motion:}For each matched feature pair $\{ \mathbf{p}_{i,j}^u, \mathbf{p}_{i,j}^v \}$, the residual inter motion ${\Delta\mathbf{m}_{i,j}^{u \to v,\text{res},inter}}$ is calculated in equ.\ref{equ:decom}.
}
\textcolor{black}{
\textbf{Vertex-Level Residual Inter Motion:}
Residual inter motions are propagated to vertices via spatial filtering:
\begin{equation}
\Delta\mathbf{n}_{i,g,h}^{{u \to v},\text{res},inter} = \sum_{j \in \mathcal{N}(g,h)} w_{gj} \Delta\mathbf{m}_{i,j}^{{u \to v},\text{res},inter}
\end{equation}
where $\mathcal{N}(g,h)$ denotes features in the ellipse of vertex $v_{g,h}$, and $w_j$ are Gaussian weights based on spatial distances.
\subsubsection{Stitching Motion Field} \label{sec:stitch_field}
The final stitching motion field combines global alignment and local adjustments:
\begin{equation}
\mathbf{n}_{i,g,h}^{{u \to v},inter} = \underbrace{\mathbf{n}_{i,g,h}^{{u \to v},\text{global},inter}}_{\text{Global}} + \underbrace{\Delta\mathbf{n}_{i,g,h}^{{u \to v},\text{res},inter}}_{\text{Residual}}
\label{eq:stitch_field}
\end{equation}
}
\textcolor{black}{
\subsection{Unified Vertex Motion and Optimization} \label{sec:unified_opt}
\subsubsection{Unified Vertex Motion Formulation} \label{sec:unified_motion}
The unified vertex motion is defined as the linear superposition of stabilization and stitching motion fields, enabling simultaneous spatiotemporal consistency in both overlapping and non-overlapping regions:
\begin{equation}
\mathbf{n}_{i,g,h}^{u,\text{unified}} = \underbrace{\mathbf{n}_{i,g,h}^{u,\text{intra}}}_{\text{Stabilization}} + \underbrace{\mathbf{n}_{i,g,h}^{{u \to v},\text{inter}}}_{\text{Stitching}}, \quad \forall i \in [1,N], \, \forall g \in [1,mn]
\label{eq:unified_motion}
\end{equation}
\textbf{Component Decomposition:}
\begin{itemize}
\item \textit{Stabilization Motion} $\mathbf{n}_{i,g,h}^{u,\text{intra}}$: Compensates intra-unit vibrations (Eq.~\ref{eq:final_motion})
\item \textit{Stitching Motion} $\mathbf{n}_{i,g,h}^{{u \to v},\text{inter}}$: Aligns inter-unit geometries (Eq.~\ref{eq:stitch_field})
\end{itemize}
\subsubsection{Vertex Motion Profiles} \label{sec:vertex_profiles}
The temporal evolution of unified motion is captured through:
\begin{itemize}
\item \textit{Accumulated Trajectory}:$T_{g,h}^u \in \mathbb{R}^{N \times 2}$: Accumulated intra-unit motion vectors simulating camera ego-motion:
\begin{equation}
T_{g,h}^u(i) = \sum_{s=1}^i \mathbf{n}_{s,g,h}^{u,\text{unified}}
\end{equation}
\item \textit{Instantaneous Profile}:$V_{g,h}^{u \to v} \in \mathbb{R}^{N \times 2}$: Inter-unit motion vectors aligning adjacent views:
\begin{equation}
V_{g,h}^{u \to v}(i) = \mathbf{n}_{i,g,h}^{u \to v,\text{inter}}
\end{equation}
\end{itemize}
}
\subsubsection{Unified Energy Formulation} \label{sec:unified_energy}
The joint optimization combines stabilization smoothness and stitching alignment:
\begin{equation}
\mathcal{L}_{\text{total}} = \underbrace{\mathcal{L}_{\text{stab}}^A + \mathcal{L}_{\text{stab}}^B}_{\text{Stabilization}} + \beta \underbrace{\mathcal{L}_{\text{stitch}}^{AB}}_{\text{Stitching}}
\label{eq:total_energy}
\end{equation}
\textcolor{black}{
\textbf{Stabilization Energy} for unit $u$:
\begin{equation}
\mathcal{L}_{\text{stab}}^u = \sum_{i=1}^{N-1} \left( \| \widehat{T}_{g,h}^u(i) - T_{g,h}^u(i) \|^2 + \lambda_t \sum_{j \in \mathcal{W}_i} \omega_{ij} \| \widehat{T}_{g,h}^u(i) - \widehat{T}_{g,h}^u(j) \|^2 \right)
\end{equation}
where $\mathcal{W}_i = [i-\sigma, i+\sigma]$ is the temporal window, $\omega_{ij} = \exp(-\frac{(j-i)^2}{( \sigma/3 )^2})$ and $\lambda_t=1.2$ balances fidelity vs smoothness. $\widehat{T}_{g,h}^u(i)$ denotes the optimized temporal motion trajectory. In other words, frames that are closer to the current frame are assigned higher weight values.
}

In the energy function $\mathcal{L}_{\text{stab}}^u$, ${\| \widehat{T}_{g,h}^u(i) - T_{g,h}^u(i) \|^2}$ serves as the preservation term of the trajectory, allowing the optimized trajectory to retain the movement trend of the original trajectory. This helps minimize cropping and distortion in the stabilized footage. $\| \widehat{T}_{g,h}^u(i) - \widehat{T}_{g,h}^u(j) \|^2$ acts as a smoothing term designed to constrain camera movement within a local temporal window to achieve consistency. This results in a smoother temporal trajectory, enhancing video stabilization. $\lambda_{t}$  serves as an adjustment factor to balance the weights of the two components.

\begin{figure}[!t]
\centering
\subfloat[Original Stitching Profiles]{\includegraphics[width=0.45\columnwidth]{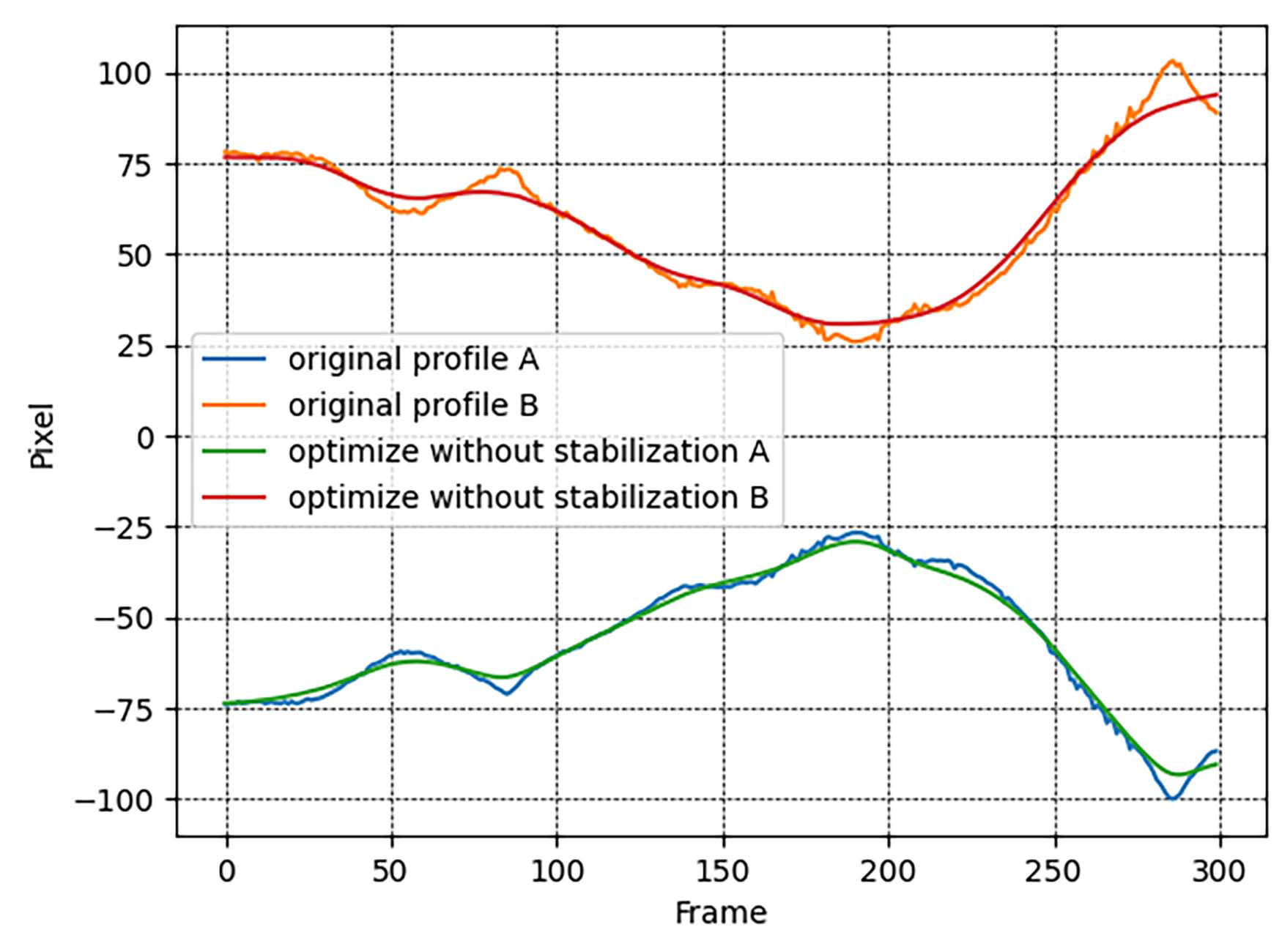}}%
\label{optimization(a)}
\hfil
\subfloat[Optimized Stitching Profiles]{\includegraphics[width=0.45\columnwidth]{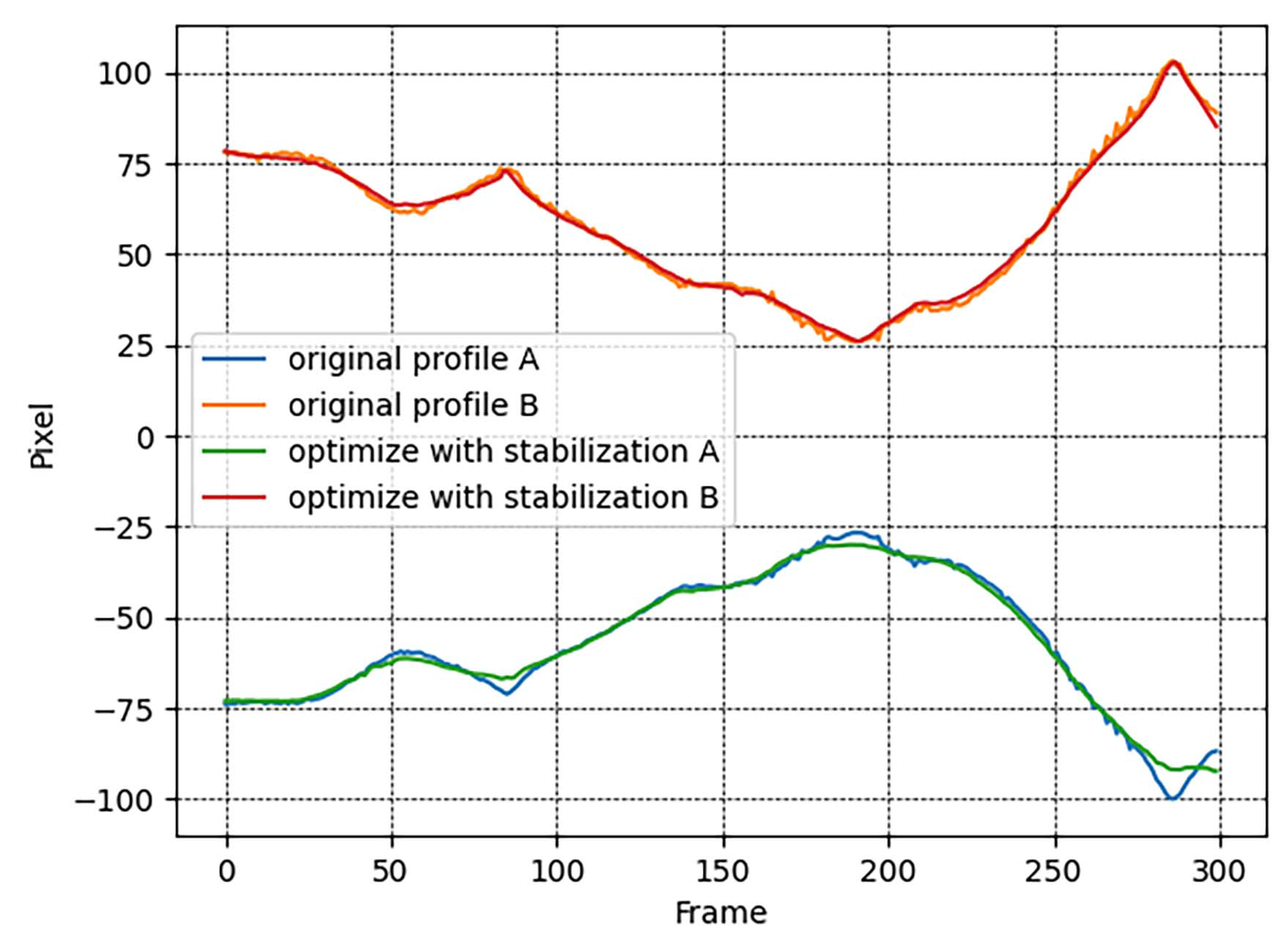}}%
\label{optimization(b)}
\hfil
  \caption{\textbf{Visualization of the Optimized Stitching Vertex Profiles:} This figure elucidates the impact of considering stabilization cost during the optimization of stitching vertex profiles. Subfigure (a) displays the original stitching profiles of camera A (blue) and camera B (orange) without considering the stabilization cost, while subfigure (b) showcases the optimized profiles (green for camera A and red for camera B) with the stabilization cost taken into account. The optimized profiles in subfigure (b) demonstrate a more harmonized alignment between the cameras, aiding in minimizing image distortions and ensuring a smoother transition in the stitched video output.}
  \label{optimization}
\end{figure}

\textcolor{black}{
\textbf{Stitching Energy} between units $A$ and $B$:
\begin{equation}
\begin{split}
\mathcal{L}_{\text{stitch}}^{AB} = \sum_{i=1}^{N}  
\underbrace{\| \widehat{V}_{g,h}^{A \to B}(i) - V_{g,h}^{A \to B}(i) + \Delta \widehat{T}_{g,h}^{A}(i) \|^2}_{\text{A→B}}\\+\sum_{i=1}^{N}\underbrace{\| \widehat{V}_{g,h}^{B \to A}(i) - V_{g,h}^{B \to A}(i) + \Delta \widehat{T}_{g,h}^{B}(i) \|^2}_{\text{B→A}} 
\end{split}
\end{equation}
}
Where $\Delta \widehat{T}_{g,h}^{A}(i) = \widehat{T}_{g,h}^A(i) - T_{g,h}^A(i) $  and $\Delta \widehat{T}_{g,h}^{B}(i) = \widehat{T}_{g,h}^B(i) - T_{g,h}^B(i)$ compensate the displacement of mesh vertices during the stabilization of input videos (as shown in Fig. \ref{optimization}).

\textcolor{black}{
\subsubsection{Optimization Strategy} \label{sec:optim}
We adopt a three-stage alternating minimization approach:
}

\textcolor{black}{
\textbf{Stage 1: Independent Stabilization}  
Solve $\min \mathcal{L}_{\text{stab}}^A$ and $\min \mathcal{L}_{\text{stab}}^B$ separately to get initial $\widehat{T}_{g,h}^A$, $\widehat{T}_{g,h}^B$.
}

\textcolor{black}{
\textbf{Stage 2: Stitching Alignment}  
Fix $\widehat{T}_{g,h}^u$, solve $\min \mathcal{L}_{\text{stitch}}^{AB}$ to update stitching profiles $\widehat{V}_{g,h}^{A \to B}$.
}

\textcolor{black}{
\textbf{Stage 3: Joint Refinement}  
Alternately update variables with:
\begin{equation}
\widehat{T}_{g,h}^u \leftarrow \arg\min \left( \mathcal{L}_{\text{stab}}^u + \beta \| \widehat{V}_{g,h}^{u \to v} - V_{g,h}^{u \to v} + \Delta \widehat{T}_{g,h}^{u} \|^2 \right)
\end{equation}
Convergence is achieved within 3 iterations in practice.
}

\subsection{Panoramic Video Generation}

After the completion of the unified optimization, the motion of each mesh vertex is computed as $\widehat{T}_{g,h}^u - {T}_{g,h}^u$ from one frame to the next. Subsequently, the images are warped based on these calculated motions. Importantly, the computation of each vertex can be executed in parallel, facilitating efficient processing.

\section{Experiments}
\subsection{Experiments Overview}
\subsubsection{Experiments Setup}
% 本实验在北京理工大学内进行验证。所有的算法均在机载平台上执行。实验主要分为两大部分，一部分是与其他SOTA方法的对比，另一部分是在我们所验证场景的自己模块的对比。
% \item 其他方法对比：我们在实际场景中采集了白天、雪景、夜间的数据集，其中case1-5是我们采集的白天数据集，case6-8是guo etal等人提供的白天数据集，case9是雪景，case10是夜间。在这些数据集上我们分别在稳定和拼接方法上进行对比。数据集的可视化切片如Fig. \ref{Fig7}所示
% \item 自身对比：为了验证在拖挂车非刚性情况下的适用性，我们采用了静态拼接的方法，用于与我们自己方法的对比。静态拼接方法标定好相机之间的相对位置参数，然后不再改变。
The experiments are primarily divided into three main parts: stabilization performance, stitching performance, and joint stabilization and stitching performance. For \textcolor{black}{a} comparative analysis with other methods, we conducted experiments using both publicly available datasets and datasets we collected ourselves.
\begin{itemize}
\item Stabilization Performance: We compared our approach with Bundled\citep{bundled}, Meshflow\citep{meshflow}, Difrint\citep{difrint}, DUT\citep{dut}, and UVSS\citep{UVSS} on our dataset. Our method achieved comparable results to the current state-of-the-art video stabilization algorithms in terms of video stability.

\item Stitching Performance: We conducted comparisons on our own dataset with AANAP\citep{AANAP}, LPC\citep{LPC}, PTGui, UDIS2\citep{UDIS2}, and UVSS\citep{UVSS}. To ensure fairness, these experiments were conducted on two-image stitching tasks. The results demonstrate that our method performs better on TTWR data.

\item Joint Stabilization and Stitching Performance: We compared our approach with \citep{joint} and \citep{dynamic} using the dataset from Guo et al.\citep{joint}. Additionally, we demonstrated the surround-view stitching results on our own dataset. The experiments indicate that our method produces a surround-view system suitable for deployment on tractor-trailer wheeled robots.
\end{itemize}

\subsubsection{Hardware Platform}
% 为了验证所提方法在实际场景中的可行性和效果，我们在一辆实验tractor-trailer vehicle上进行了实验，其由两部分组成，首先是车头tractor部分，其由上汽通用五菱宝骏E100作为主体，然后在其后方固定了铰接装置，trailer部分由手工架与轮子连接而成。这个tractor-trailer系统总长为4.5m, 宽为1.5m, 高为1.8m。该系统搭载了Nvidia Jetson AGX Orin作为系统处理器，连接了八个相机进行数据采集，其中三个相机固连在tractor上，用于感知前方、左前、右前的视场，五个相机固连在trailer上，用于感知左方、左后、后方、右后、右方的视场。考虑到时间和计算成本，单个相机的视场角为120度，采集的图像为1280*720。具体的实验设备为图所示。
To validate the feasibility and efficacy of the proposed method in real-world scenarios, we conducted experiments on a practical TTWR. This robot consists of two main components: the front tractor unit and the trailing part. The tractor unit is primarily based on the SAIC-GM-Wuling Baojun E100. An articulated hitch was affixed to its rear, which connected it to the trailer. The trailer itself was manually constructed, consisting of a frame connected to wheels. The overall dimensions of this TTWR are 4.5 meters in length, 1.5 meters in width, and 1.8 meters in height.

\begin{figure}[!t]
\centering
\includegraphics[width=0.5\linewidth]{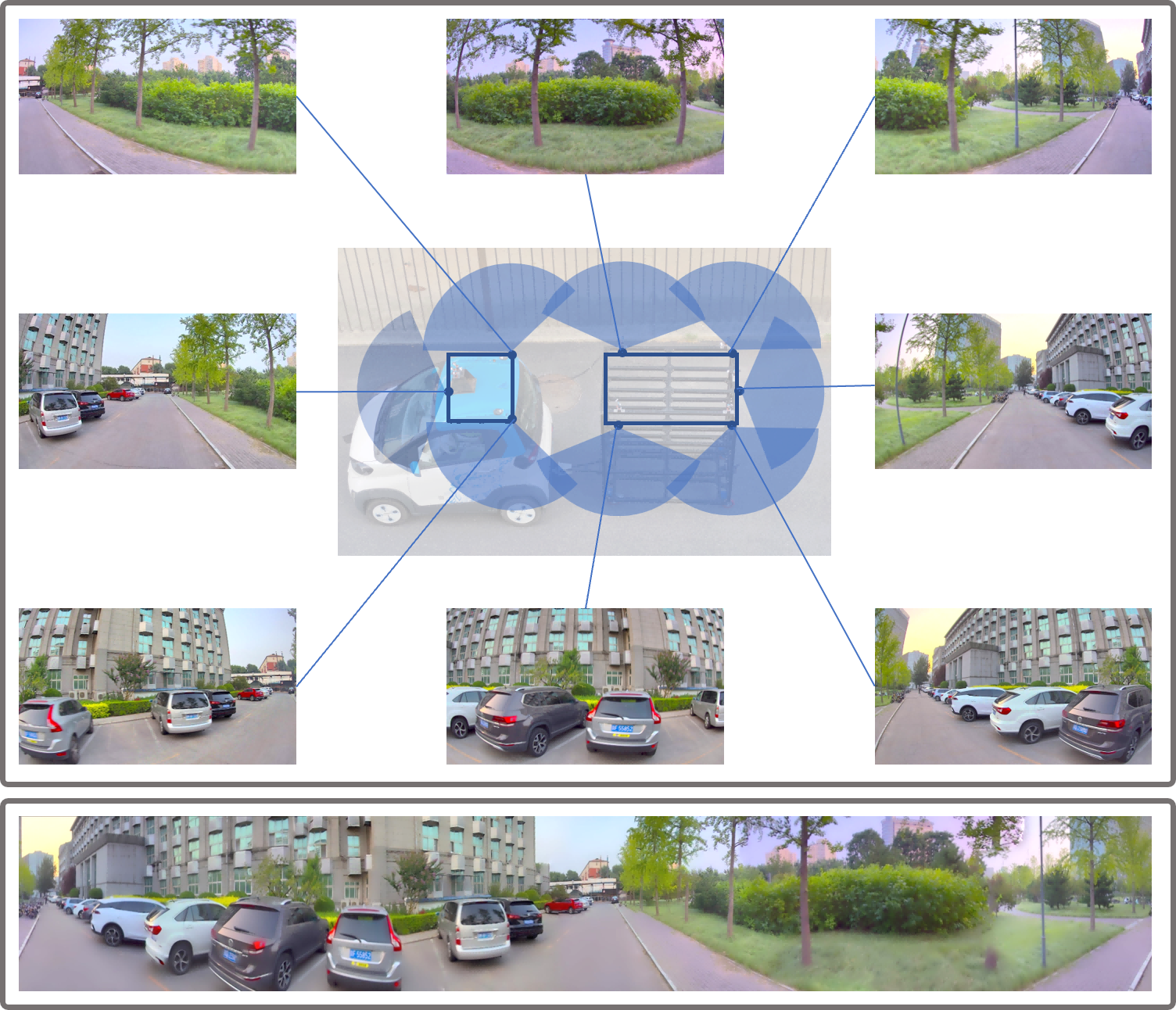}
\caption{\textbf{Experimental Hardware Setup:} Depicting the TTWR utilized for real-world experimentation, showcasing the distinct placement of cameras and the articulated hitch connecting the tractor and trailer units. The tractor unit, primarily based on the SAIC-GM-Wuling Baojun E100, along with the manually constructed trailer, forms the basis of the experiment. The camera arrangements for comprehensive field-of-view coverage are also highlighted.}
\label{fig:hardware}
\end{figure}

The system is equipped with an Nvidia Jetson AGX Orin as its central processor. Eight cameras are integrated into the system for data acquisition. Of these, three cameras are rigidly mounted on the tractor to perceive the forward, left-front, and right-front fields of view. The remaining five cameras are affixed to the trailer and dedicated to capturing the left, left-rear, rear, right-rear, and right fields of view. Given considerations of time and computational cost, each camera possesses a field of view of 120 degrees and captures images at a resolution of 1280 x 720 pixels. A detailed depiction of the experimental apparatus is illustrated in Fig. \ref{fig:hardware}.

\subsubsection{Datasets}
% 我们在七个case下进行了我们的实验，他们都基于同一地理场景下，其中Case 1是由手持智能手机捕获的，而case 2是从数据采集车辆拍摄的视频中截取的，case3-5分别是拖挂车系统的前方视图（由前相机拍摄），侧方视图（右相机拍摄），后方视图（后相机拍摄）case9是采集的雪景视图，case10是采集的夜间数据。在此场景下，我们的方法与其他方法产生的结果如表1所示。此外，我们还对case2场景的稳定过程进行了可视化，如图所示。
We made our own dataset across seven cases, all based on the same geographic scenario. Case 1 was captured using a handheld smartphone, while Case 2 was extracted from video footage recorded by a data collection robot. Cases 3-5 represent different views from a trailer system: the front view (captured by the front camera), the side view (captured by the right camera), and the rear view (captured by the rear camera). Case 6 includes snow scene views, and Case 7 comprises nighttime data.

%我们在公开数据集\citep{joint}上进行Joint Stabilization and Stitching Performance的对比实验。case8-10分别是数据集中的经典场景。 Visualization slices of the datasets are shown in Fig. \ref{fig:case}.

\begin{figure}[!t]
\centering
\includegraphics[width=0.5\linewidth]{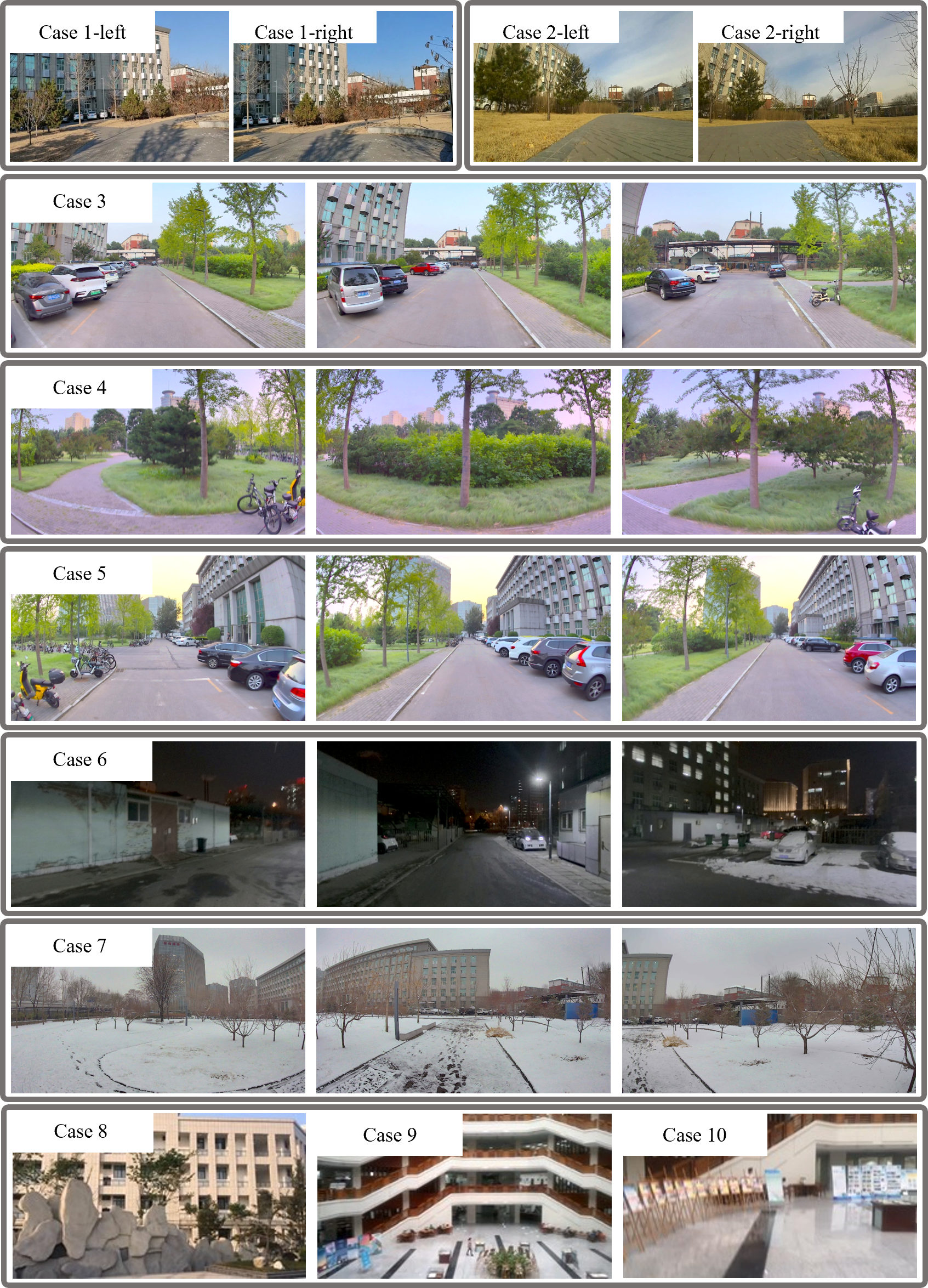}
\caption{\textbf{Visualization of the experimental datasets for cases 1-10:} This visualization provides a brief overview of our experimental scenarios. Cases 1-5 represent daytime scenes collected by us. Cases 6-8 are daytime datasets provided by Guo et al. Case 9 comprises snow scene data collected at the Beijing Institute of Technology in December 2023, while Case 10 includes nighttime data collected during the same period.}
\label{fig:case}
\end{figure}

We conducted comparative experiments on Joint Stabilization and Stitching Performance using the public dataset \citep{joint}. Cases 8-10 represent classic scenarios within the dataset. Visualization slices of these datasets are presented in Fig. \ref{fig:case}.

\subsection{Results}
\subsubsection{Stabilization Performance}
% 我们采用文献Bundled camera paths for video stabilization中提出的客观定量评价体系，其中包括cropping ratio and global distortion和stability。简单来说，cropping ratio衡量图形变形后仍保留的原图像比例；global distortion衡量图形失真程度，如文献中所说，这里的程度应为整个视频每帧图像计算后的最小值；stability是防抖效果最直观的部分，如文献中所说，其评价的是运动低频部分所包含的能量。根据评价标准，这三个值越接近1，则效果越好。
We adopted the objective quantitative evaluation system proposed in \citep{bundled}, which encompasses the metrics of cropping ratio, global distortion, and stability. Briefly, the cropping ratio measures the proportion of the original image retained after geometric deformation; global distortion evaluates the degree of distortion, where, as mentioned in the literature, the degree should be the minimum value computed across all frames of the entire video; stability reflects the anti-shake effect most intuitively, as the literature articulates, it assesses the energy contained in the low-frequency motion part. According to the evaluation criteria, the closer these three values approach 1, the better the performance.

\begin{figure}[!t]
\centering

\subfloat[ The curves before and after stabilization]{\includegraphics[width=0.6\linewidth]{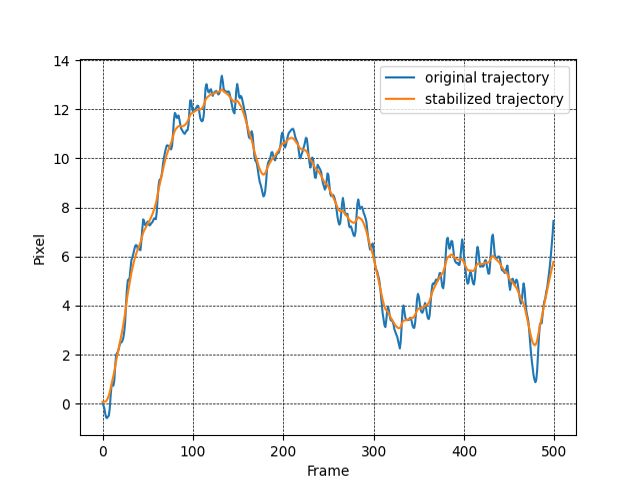}}
\hfil
\subfloat[ The visual effect of stabilization]{\includegraphics[width=0.3\linewidth]{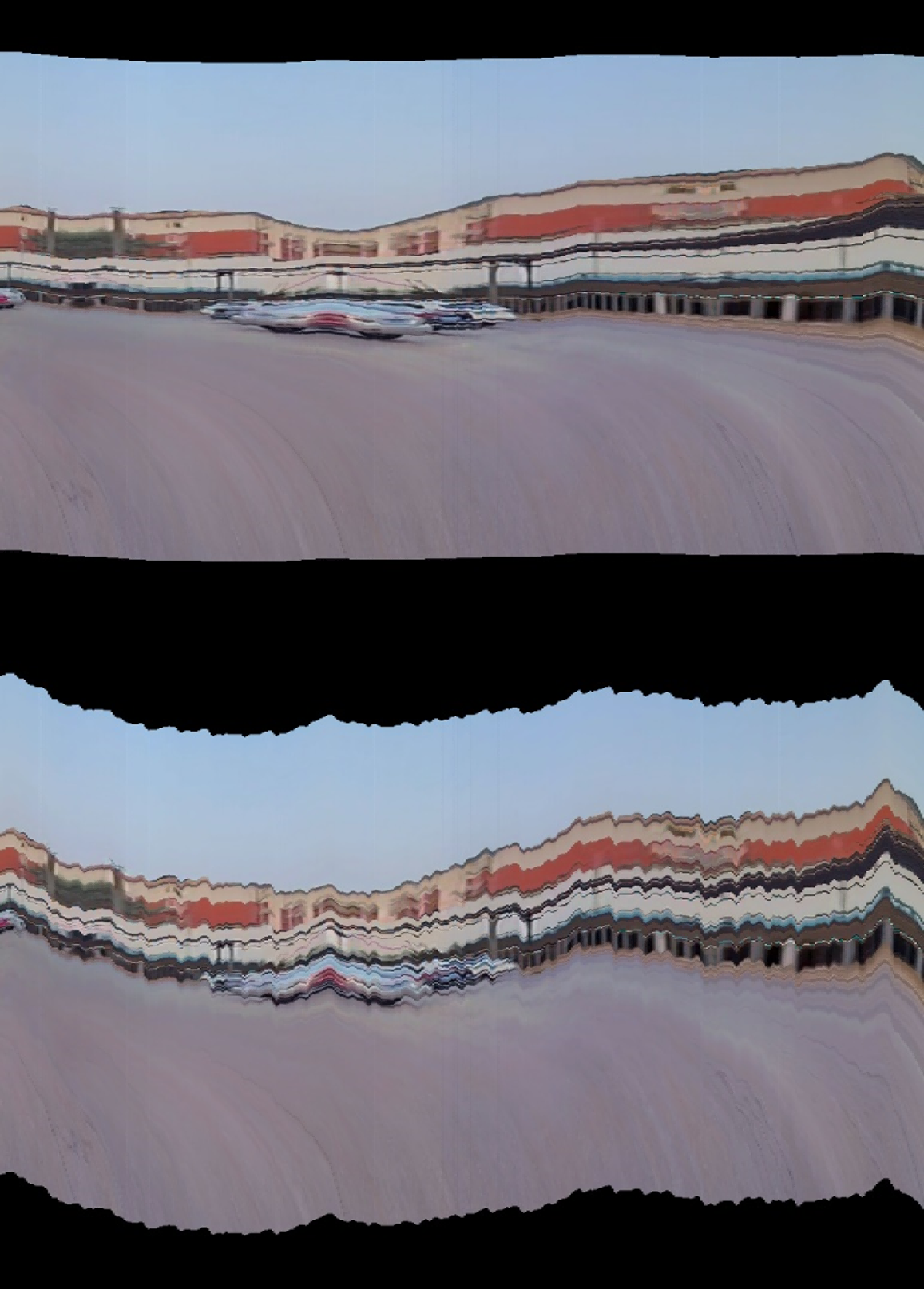}}

\caption{\textbf{Stabilization Visualization}: Subfigure (a) illustrates the curves before and after smoothing, demonstrating the effectiveness of our stabilization process. Subfigure (b) provides the visual effect of this smoothing on the recorded footage, highlighting the enhancements in image stability for the forward view captured by the trailer system's front camera.}
\label{fig:curves}
\end{figure}

The outcomes generated by our method in comparison with other methods in this scenario are illustrated in Table \ref{tab:stablization_table}. Additionally, we visualized the stabilization process of the scenario in Case 3, as depicted in Fig. \ref{fig:curves}.

\begin{table}[htbp]
% \renewcommand\arraystretch{1.5}
% \setlength{\abovecaptionskip}{0cm}  % 段前
% \setlength{\belowcaptionskip}{0cm} % 段后
% \vspace{10pt}  % 添加10pt的额外距离
\caption{Comparison with previous stabilization approaches}
\label{tab:stablization_table}
\begin{center}
% \begin{tabular}{c|c|p{1.5cm}|p{1.3cm}|p{1.5cm}|p{1.3cm}|p{0.8cm}|p{0.8cm}|p{0.8cm}|p{0.8cm}|p{0.8cm}|p{0.8cm}}
\resizebox{\linewidth}{!}{
\begin{tabular}{c|c|c|c|c|c|c|c|c|c|c|c}
    \hline
    Metrics & Methods & Case1-Right & Case1-Left & Case2-Right & Case2-Left & Case3 & Case4 & Case5 & Case9 & Case10 & Average \\
    \hline
    \multirow{6}{*}{Cropping} 
                              & Bundled\citep{bundled}  & \multicolumn{1}{c|}{0.9801} & \multicolumn{1}{c|}{0.9920} & \multicolumn{1}{c|}{0.9453} & \multicolumn{1}{c|}{0.9138} & \multicolumn{1}{c|}{0.9914} & \multicolumn{1}{c|}{0.8861} & \multicolumn{1}{c|}{\textbf{0.9705}} & \multicolumn{1}{c|}{0.8497} & \multicolumn{1}{c|}{0.8466}& \multicolumn{1}{c}{0.9306}\\
                              & Meshflow\citep{meshflow} & \multicolumn{1}{c|}{0.9284} & \multicolumn{1}{c|}{0.9636} & \multicolumn{1}{c|}{0.9031} & \multicolumn{1}{c|}{0.8548} & \multicolumn{1}{c|}{0.8682} & \multicolumn{1}{c|}{0.8866} & \multicolumn{1}{c|}{0.8732} & \multicolumn{1}{c|}{0.9312} & \multicolumn{1}{c|}{0.8428}  & \multicolumn{1}{c}{0.8947}\\
                              & Difrint\citep{difrint}  & \multicolumn{1}{c|}{0.9705} & \multicolumn{1}{c|}{0.9731} & \multicolumn{1}{c|}{0.9460} & \multicolumn{1}{c|}{0.8896} & \multicolumn{1}{c|}{0.9713} & \multicolumn{1}{c|}{0.9514} & \multicolumn{1}{c|}{0.9682} & \multicolumn{1}{c|}{0.9828} & \multicolumn{1}{c|}{0.9752}
                              & \multicolumn{1}{c}{0.9587}\\
                              & DUT\citep{dut}      & \multicolumn{1}{c|}{0.8855} & \multicolumn{1}{c|}{0.7864} & \multicolumn{1}{c|}{0.9653} & \multicolumn{1}{c|}{0.7556} & \multicolumn{1}{c|}{0.9882} & \multicolumn{1}{c|}{0.8072} & \multicolumn{1}{c|}{0.7317} & \multicolumn{1}{c|}{0.7525} & \multicolumn{1}{c|}{0.7724} 
                              & \multicolumn{1}{c}{0.8272}\\
                              & UVSS\citep{UVSS}     & \multicolumn{1}{c|}{0.9306} & \multicolumn{1}{c|}{0.9650} & \multicolumn{1}{c|}{0.9130} & \multicolumn{1}{c|}{0.8534} & \multicolumn{1}{c|}{0.8729} & \multicolumn{1}{c|}{0.8949} & \multicolumn{1}{c|}{0.9053} & \multicolumn{1}{c|}{0.8919} & \multicolumn{1}{c|}{0.9056}
                              & \multicolumn{1}{c}{0.9036}\\
                              & Ours     & \multicolumn{1}{c|}{\textbf{0.9999}} & \multicolumn{1}{c|}{\textbf{0.9996}} & \multicolumn{1}{c|}{\textbf{0.9909}} & \multicolumn{1}{c|}{\textbf{0.9960}} & \multicolumn{1}{c|}{\textbf{1.000}} & \multicolumn{1}{c|}{\textbf{0.9830}} & \multicolumn{1}{c|}{0.8886} & \multicolumn{1}{c|}{\textbf{0.9885}} & \multicolumn{1}{c|}{\textbf{0.9990}} 
                              & \multicolumn{1}{c}{\textbf{0.9828}}\\
    \hline
    \multirow{6}{*}{Distortion} 
                                & Bundled\citep{bundled}  & \multicolumn{1}{c|}{\textbf{0.9818}} & \multicolumn{1}{c|}{0.9929} & \multicolumn{1}{c|}{0.9656} & \multicolumn{1}{c|}{0.9248} & \multicolumn{1}{c|}{0.9563} & \multicolumn{1}{c|}{0.8900} & \multicolumn{1}{c|}{0.9675} & \multicolumn{1}{c|}{0.8449} & \multicolumn{1}{c|}{0.8324} & \multicolumn{1}{c}{0.9285}\\
                                & Meshflow\citep{meshflow} & \multicolumn{1}{c|}{0.9628} & \multicolumn{1}{c|}{0.9919} & \multicolumn{1}{c|}{0.9481} & \multicolumn{1}{c|}{0.9455} & \multicolumn{1}{c|}{0.9595} & \multicolumn{1}{c|}{0.9630} & \multicolumn{1}{c|}{0.9284} & \multicolumn{1}{c|}{0.9712} & \multicolumn{1}{c|}{0.8992} & \multicolumn{1}{c}{0.9522}\\
                                & Difrint\citep{difrint}  & \multicolumn{1}{c|}{0.9663} & \multicolumn{1}{c|}{0.9711} & \multicolumn{1}{c|}{0.9416} & \multicolumn{1}{c|}{0.9546} & \multicolumn{1}{c|}{0.9661} & \multicolumn{1}{c|}{0.9585} & \multicolumn{1}{c|}{0.9736} & \multicolumn{1}{c|}{0.9648} & \multicolumn{1}{c|}{0.9425} 
                                & \multicolumn{1}{c}{0.9599}\\
                                & DUT\citep{dut}      & \multicolumn{1}{c|}{0.7853} & \multicolumn{1}{c|}{0.8346} & \multicolumn{1}{c|}{0.7490} & \multicolumn{1}{c|}{0.7332} & \multicolumn{1}{c|}{0.5904} & \multicolumn{1}{c|}{0.6514} & \multicolumn{1}{c|}{0.7679} & \multicolumn{1}{c|}{0.7621} & \multicolumn{1}{c|}{0.7737} 
                                & \multicolumn{1}{c}{0.7386}\\
                                & UVSS\citep{UVSS}     & \multicolumn{1}{c|}{0.9621} & \multicolumn{1}{c|}{0.9922} & \multicolumn{1}{c|}{0.9693} & \multicolumn{1}{c|}{0.9444} & \multicolumn{1}{c|}{0.9591} & \multicolumn{1}{c|}{0.9502} & \multicolumn{1}{c|}{0.9644} & \multicolumn{1}{c|}{0.9715} & \multicolumn{1}{c|}{0.8340} 
                                & \multicolumn{1}{c}{0.9497}\\
                                & Ours     & \multicolumn{1}{c|}{0.9613} & \multicolumn{1}{c|}{\textbf{0.9936}} & \multicolumn{1}{c|}{\textbf{0.9707}} & \multicolumn{1}{c|}{\textbf{0.9742}} & \multicolumn{1}{c|}{\textbf{0.9902}} & \multicolumn{1}{c|}{\textbf{0.9789}} & \multicolumn{1}{c|}{\textbf{0.9742}} & \multicolumn{1}{c|}{\textbf{0.9905}} & \multicolumn{1}{c|}{\textbf{0.9907}} 
                                & \multicolumn{1}{c}{\textbf{0.9805}}\\
    \hline
    \multirow{6}{*}{Stability}  & Bundled\citep{bundled}  & \multicolumn{1}{c|}{0.8627} & \multicolumn{1}{c|}{0.9376} & \multicolumn{1}{c|}{0.9341} & \multicolumn{1}{c|}{0.8783} & \multicolumn{1}{c|}{0.8569} & \multicolumn{1}{c|}{0.5664} & \multicolumn{1}{c|}{0.9393} & \multicolumn{1}{c|}{0.8554} & \multicolumn{1}{c|}{0.8991} & \multicolumn{1}{c}{0.8589}\\
                                & Meshflow\citep{meshflow} & \multicolumn{1}{c|}{0.8293} & \multicolumn{1}{c|}{0.7603} & \multicolumn{1}{c|}{0.5246} & \multicolumn{1}{c|}{0.4527} & \multicolumn{1}{c|}{0.1282} & \multicolumn{1}{c|}{0.0942} & \multicolumn{1}{c|}{0.1410} & \multicolumn{1}{c|}{0.0662} & \multicolumn{1}{c|}{0.1885} & \multicolumn{1}{c}{0.3539}\\
                                & Difrint\citep{difrint}  & \multicolumn{1}{c|}{0.8780} & \multicolumn{1}{c|}{0.9471} & \multicolumn{1}{c|}{0.9097} & \multicolumn{1}{c|}{0.8830} & \multicolumn{1}{c|}{0.8408} & \multicolumn{1}{c|}{0.6677} & \multicolumn{1}{c|}{0.9315} & \multicolumn{1}{c|}{0.8800} & \multicolumn{1}{c|}{0.9017} 
                                & \multicolumn{1}{c}{0.8711}\\
                                & DUT\citep{dut}      & \multicolumn{1}{c|}{0.8657} & \multicolumn{1}{c|}{0.7935} & \multicolumn{1}{c|}{0.7581} & \multicolumn{1}{c|}{0.6517} & \multicolumn{1}{c|}{0.1700} & \multicolumn{1}{c|}{0.1560} & \multicolumn{1}{c|}{0.1884} & \multicolumn{1}{c|}{0.1790} & \multicolumn{1}{c|}{0.4369} 
                                & \multicolumn{1}{c}{0.4666}\\
                                & UVSS\citep{UVSS}     & \multicolumn{1}{c|}{0.9342} & \multicolumn{1}{c|}{0.9403} & \multicolumn{1}{c|}{0.9342} & \multicolumn{1}{c|}{0.9038} & \multicolumn{1}{c|}{0.8587} & \multicolumn{1}{c|}{0.7782} & \multicolumn{1}{c|}{0.9348} & \multicolumn{1}{c|}{0.8897} & \multicolumn{1}{c|}{0.9037} 
                                & \multicolumn{1}{c}{0.8975}\\
                                & Ours     & \multicolumn{1}{c|}{\textbf{0.9352}} & \multicolumn{1}{c|}{\textbf{0.9839}} & \multicolumn{1}{c|}{\textbf{0.9353}} & \multicolumn{1}{c|}{\textbf{0.9060}} & \multicolumn{1}{c|}{\textbf{0.8748}} & \multicolumn{1}{c|}{\textbf{0.8275}} & \multicolumn{1}{c|}{\textbf{0.9427}} & \multicolumn{1}{c|}{\textbf{0.8913}} & \multicolumn{1}{c|}{\textbf{0.9123}} & \multicolumn{1}{c}{\textbf{0.9121}}\\
    \hline
\end{tabular}
}
\end{center}
\end{table}

%在实验中，我们发现cropping ratio这一项指标通过我们的方法在七个case上都达到很高的水平，这意味着图像在处理以后的信息损失很少。我们认为这是因为unified vertex motion model在生成时引入了空间相邻图像的信息，（正如前面所示，他是stitching motion field 和 stabilization motion field的集合），再图像处理以后，相邻空间的信息补充了因为防抖所导致的信息损失，这也是环视系统的优势所在。同理在distortion上，我们的方法同时接受stitch和stabilizaiotn的约束，因此在失真度上有更好的性能。

In our experiments, we discovered that our method achieved high cropping ratio levels across all seven cases, indicating minimal information loss after image processing. We attribute this to the incorporation of spatially adjacent image information when generating the unified vertex motion model (as previously shown, it is a combination of the stitching motion field and the stabilization motion field). After image processing, the information from adjacent spatial areas compensates for the information loss caused by stabilization, which is a significant advantage of the surround-view system. Similarly, in terms of distortion, our method outperforms others because it simultaneously adheres to both stitching and stabilization constraints, resulting in superior performance in minimizing distortion.

% 在实验中，我们认为评价低于0.4则直接失败，不再与其他方法进行比较。在大面积的天空等无纹理区域上，我们的方法通过鲁棒性特征取得了比传统方法更优异的结果。得益于我们方法对拼接与稳定的高度归一化，使得其在最终处理时没有过多损害稳定性能，在视频稳定性方面与当前最先进的视频稳定算法取得了相似的结果。

% In the experiments, we considered a score below 0.4 as a direct failure, and no further comparison with other methods was conducted.
In large untextured areas, such as expanses of sky, our method achieved superior results compared to traditional approaches, owing to the robust feature extraction employed. Thanks to the high normalization of stitching and stabilization in our method, the final processing does not overly compromise stabilization performance, achieving results on par with the current state-of-the-art video stabilization algorithms in terms of video stability.

\subsubsection{Stitching Performance}

\begin{figure}[!t]
\centering
\includegraphics[width=\linewidth]{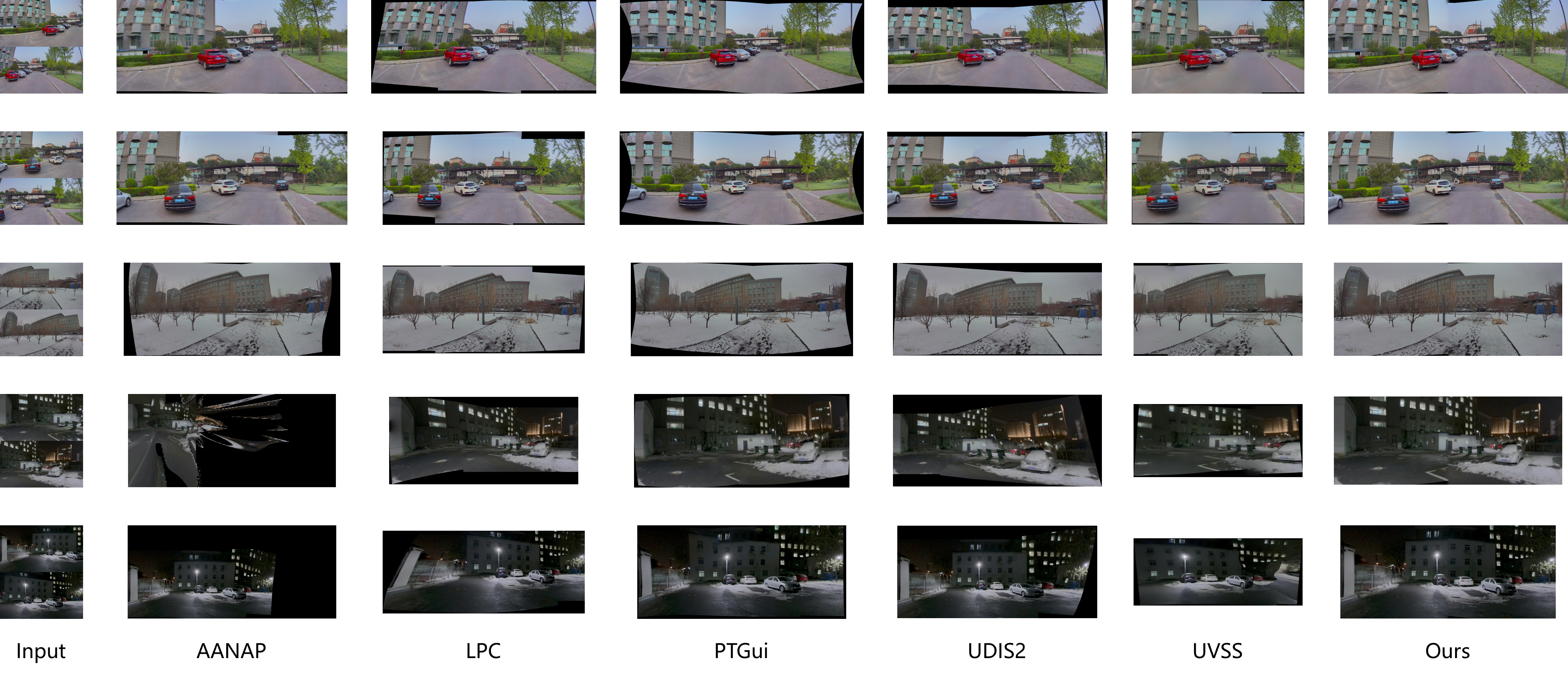}
\caption{\textbf{Comparison with previous stitching approaches on our Dataset:} In the "Input" column, the top of each row represents the left view, while the bottom represents the right view. From top to bottom, the first and second rows are daytime data, the third row is snow scene data, and the fourth and fifth rows are nighttime data. The comparisons are made with AANAP\citep{AANAP}, LPC\citep{LPC}, PTGui, UDIS2\citep{UDIS2}, and UVSS\citep{UVSS}.}
\label{fig:Figstitch}
\end{figure}

%接着我们在自己的数据集上对比我们的方法与目前sota方法UDIS2、UVSS、LPC，传统方法AANAP和专业拼接软件PTGui(https://ptgui.com)之间在两图拼接上的拼接效果。数据来源于前面我们自己采集的白天、雪景、夜间数据。最终的实验结果如图\ref{Figstitch}所示。
%在实验的过程中，LPC算法由于涉及到线特征的处理，再提取失败的情况下我们手动增加了一些线特征。PTGui软件在使用时涉及一些相机模型的选择和参数的选择。我们都调整其参数以尽可能获得较好的结果。
%从实验结果可以看出，AANAP，LPC，UDIS2的拼接逻辑都是将一张图投影到另一种图上进行拼接，在面向拖挂车这种宽基线的数据时往往会在最后的结果图边缘产生黑影或较大扭曲。PTGui，UVSS和我们的方法都是同时对图像进行扭曲，但由于PTGui需要对模型参数进行手动调整，因此需要耗费大量时间,UVSS在左右有很明显的信息损失。综合来看，我们的方法在拖挂车环视数据上取得了更好的效果。

We compare our method with the current state-of-the-art methods, UDIS2\citep{UDIS2}, UVSS\citep{UVSS}, and LPC\citep{LPC}, as well as traditional methods like AANAP\citep{AANAP} and the professional stitching software PTGui (https://ptgui.com) on our dataset. The data includes daytime, snow scene, and nighttime imagery that we collected ourselves. The final experimental results are shown in Fig. \ref{fig:Figstitch}.

During the experiments, we manually added some line features to the LPC algorithm when it failed to extract them. For the PTGui software, we adjusted the camera model selection and parameters to achieve the best possible results. 
The experimental findings reveal that the stitching algorithms of AANAP, LPC, and UDIS2 primarily function by overlaying one image onto another. This approach frequently leads to the appearance of black shadows or pronounced distortions along the edges of the resultant stitched image, particularly in scenarios involving wide-baseline data such as those from tractor-trailer surround views. In contrast, PTGui, UVSS, and our technique all employ a simultaneous warping of images. However, PTGui necessitates manual tweaking of model parameters, a process that proves to be labor-intensive, while UVSS tends to suffer from a noticeable loss of information on the peripheral left and right sides. Overall, our method distinctly outperforms others, especially in handling the complex data associated with TTWRs, achieving superior stitching quality and consistency.

\subsubsection{Joint Stabilization and Stitching Performance}

\begin{figure}[!t]
\centering
\subfloat[results of \citep{joint}]{\includegraphics[width=0.31\columnwidth]{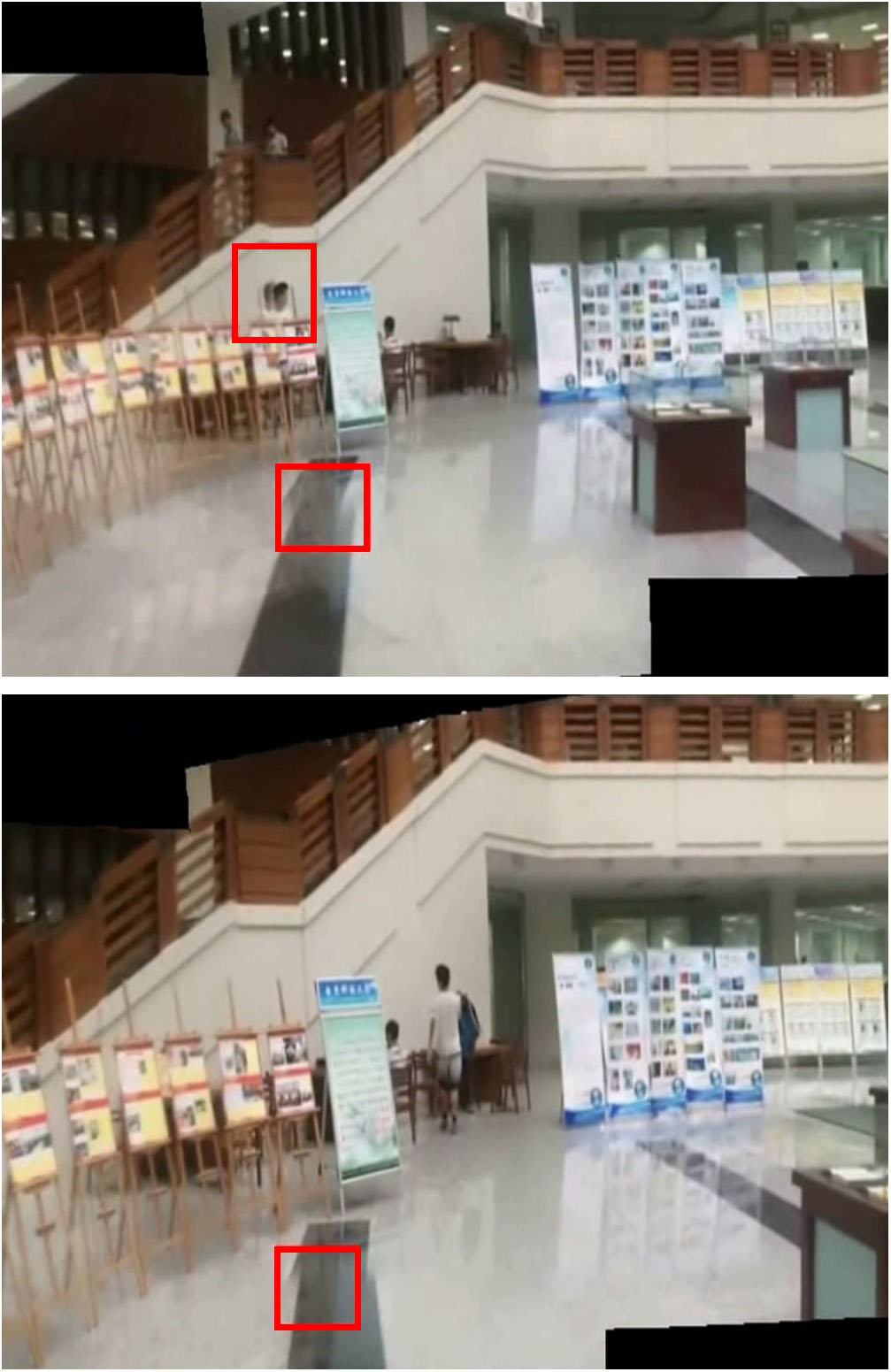}}%
\label{compare_joint}\hspace{-3pt}
\hfil
\subfloat[results of \citep{dynamic}]{\includegraphics[width=0.31\columnwidth]{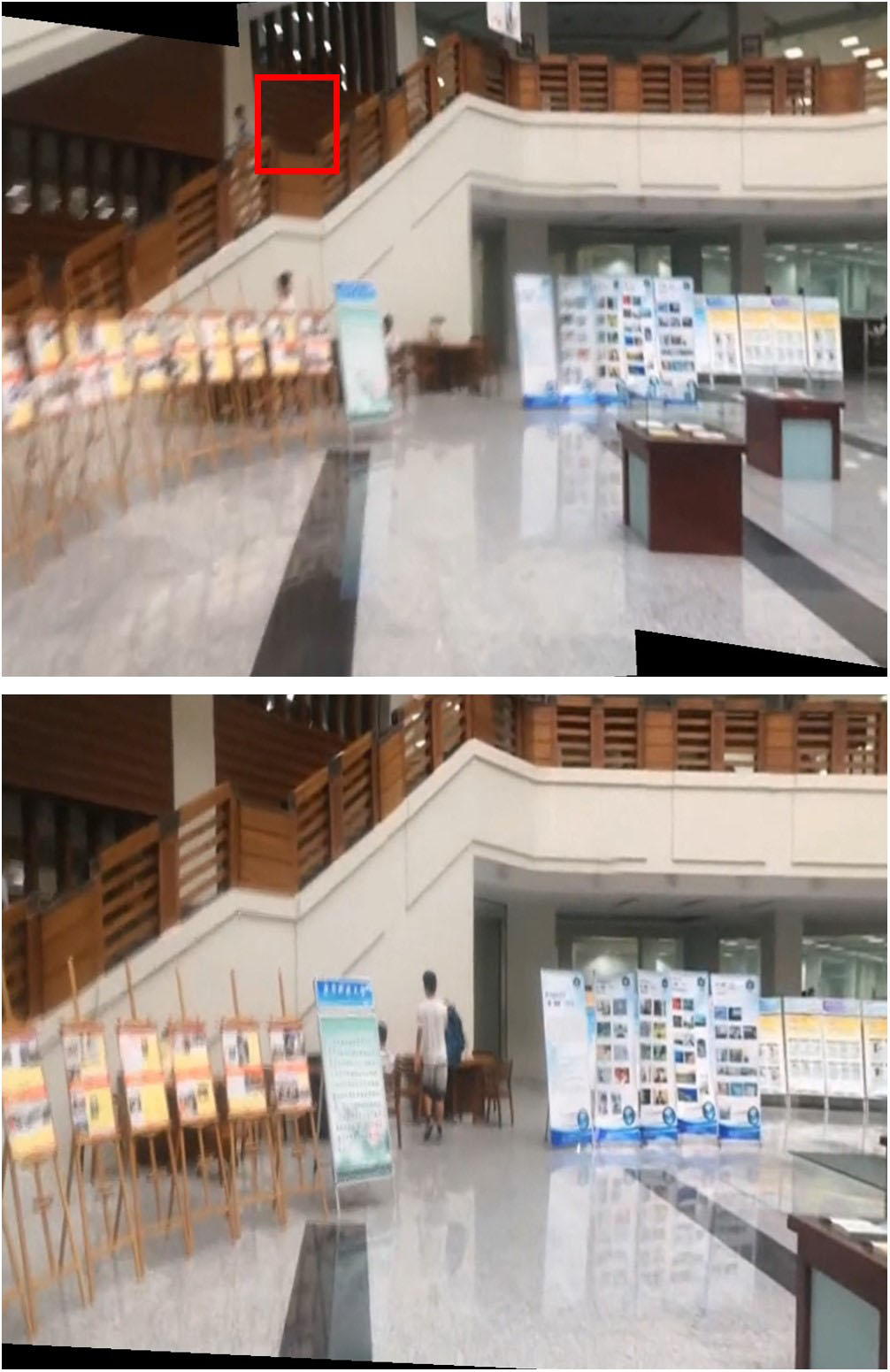}}%
\label{compare_dynamic}\hspace{-3pt}
\hfil
\subfloat[ours]{\includegraphics[width=0.31\columnwidth]{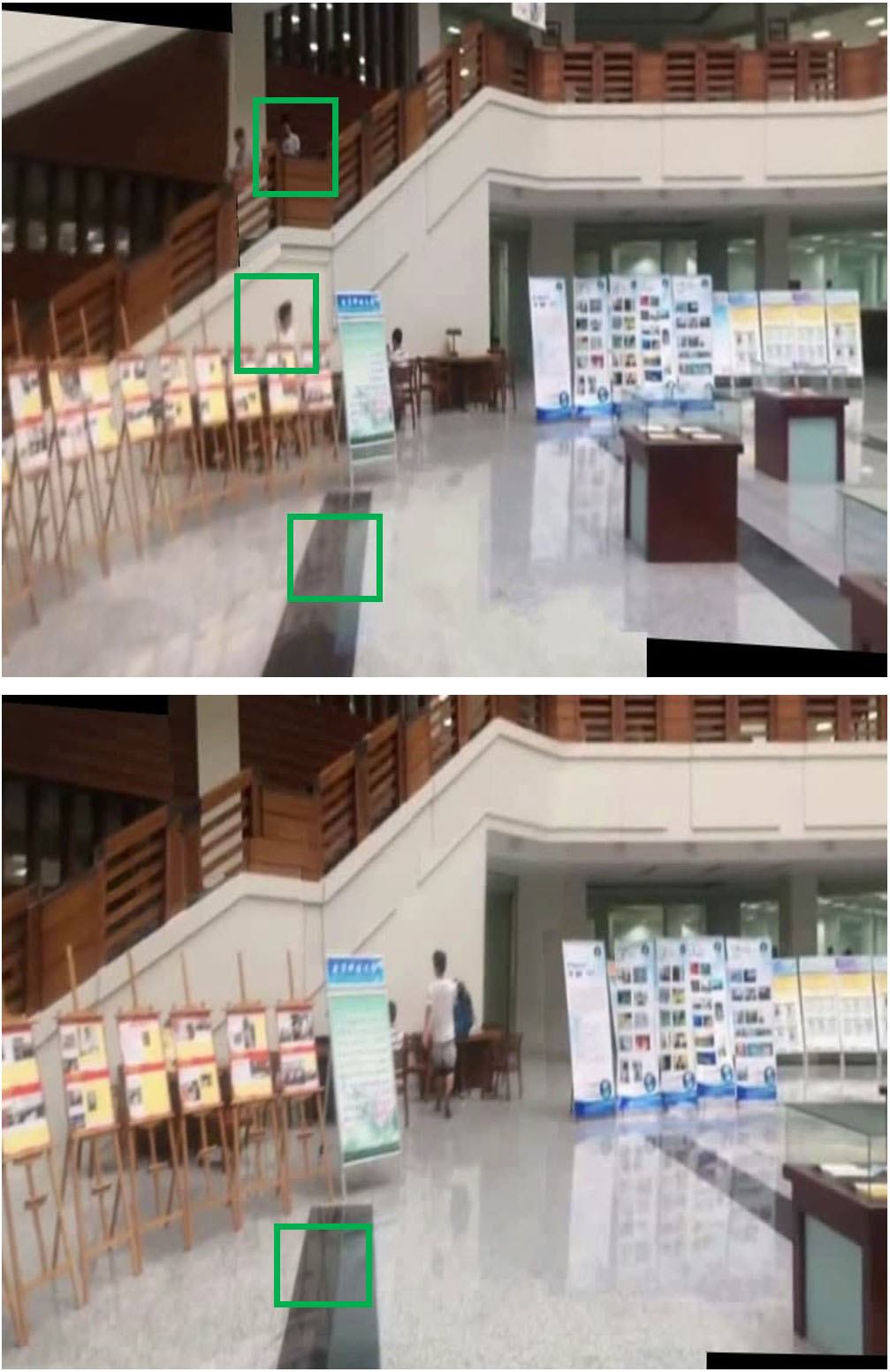}}%
\label{compare_ours}
\caption{\textbf{Subjective comparison with \citep{joint} and \citep{dynamic} in case 7.} Left to right: results of \citep{joint}, results of \citep{dynamic}, and ours.}
  \label{fig:stitching_compare}
\end{figure}

We utilize the score proposed in  Guo et al.\citep{joint} for the quantitative evaluation of outcomes and assess algorithm efficiency by measuring the computation time per frame. The score, as defined in  Guo et al.\citep{joint}, serves as a quantitative metric for evaluating the quality of results. It is computed based on the reprojection error of feature points, where a lower score signifies more accurate stitching. Additionally, to provide a more intuitive assessment of method performance, we conduct subjective comparisons. The stitching score for a single frame is determined by the average error across all feature pairs, and the highest value among all frames is designated as the final score.

We utilized a framework akin to the one presented in  Guo et al.\citep{joint} and  Nie et al.\citep{dynamic} to concurrently achieve video stabilization and stitching within a unified framework. The distinction lies in the fact that all of our optimizations are executed on mesh vertices and the motion vectors associated with each vertex, as opposed to directly regressing the homography for each mesh. For comparison with  Guo et al.\citep{joint} and  Nie et al.\citep{dynamic}, we carried out experiments on the dataset proposed by Guo et al.\citep{joint}. The video resolutions were uniformly resized to 540P (${960}\times{540}$).

\begin{table}[htbp]
\renewcommand\arraystretch{1.5}
\setlength{\abovecaptionskip}{0cm}  % 段前
\setlength{\belowcaptionskip}{-0.5cm} % 段后
\vspace{10pt}  % 添加10pt的额外距离
\caption{Comparison with previous stitching approaches on the Guo et al. dataset}
\label{tab:stitching_table}
\begin{center}
\resizebox{0.5\linewidth}{!}{
\begin{tabular}{c|c|c c c}
    \hline
    Metrics & Methods & Case 5 & Case 6 & Case 7 \\
    % \hline
    % \multirow{3}{*}{Stability} & \cite{joint} &0.892 &0.9 &0.797 \\
    %          & \cite{dynamic} &0.923 &0.923 &0.862 \\
    %          & Ours &0.866 &\textbf{0.905} &0.767 \\
    \hline
    \multirow{3}{*}{Stitching Score} & \citep{joint} &1.04 &1.17 &- \\
             & \citep{dynamic} &1.81 &2.73 &2.24 \\
             & Ours &\textbf{0.952} &\textbf{0.946} &\textbf{0.859} \\
    \hline
    \multirow{3}{*}{\makecell[c]{Computation Time\\(s/frame)}} & \citep{joint} &- &- &- \\
             & \citep{dynamic} &8.16 &7.13 &8.53 \\
             & Ours &\textbf{1.44} &\textbf{1.53} &\textbf{1.97} \\
    \hline
\end{tabular}
}
\end{center}
\end{table}
   
Table \ref{tab:stitching_table} presents the stitching score and computational time for the outcomes. Our approach effectively addresses all scenarios in cases 8 to 10, demonstrating superior performance compared to \citep{joint}, which experiences failure in case 10. The subjective comparison is visualized in Fig. \ref{fig:stitching_compare}. Notably, our method achieves slightly higher stitching scores compared to both \citep{joint} and \citep{dynamic}. Nevertheless, it is imperative to acknowledge that the stitching score is quantified in pixels, and experimental findings indicate that scores below 10 exert negligible influence on human perception following the multi-band blending of stitched images. It is of significance that our method substantially reduces computation time in comparison to \citep{dynamic} while yielding comparable results.

\begin{figure}[!t]
\centering
\subfloat[]{\includegraphics[width=\linewidth]{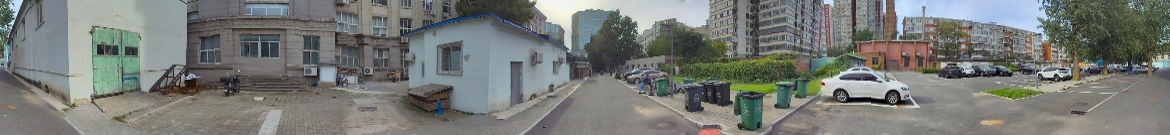}}
\hfil
\subfloat[]{\includegraphics[width=\linewidth]{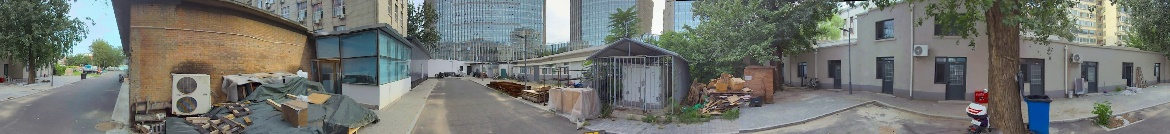}}
\hfil
\subfloat[]{\includegraphics[width=\linewidth]{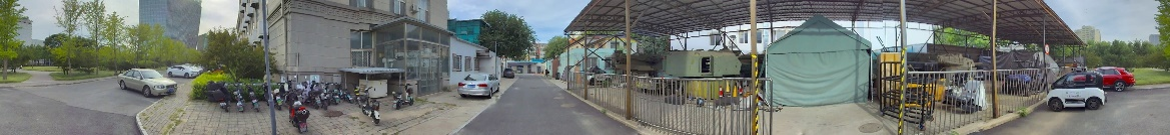}}
\hfil
\subfloat[]{\includegraphics[width=\linewidth]{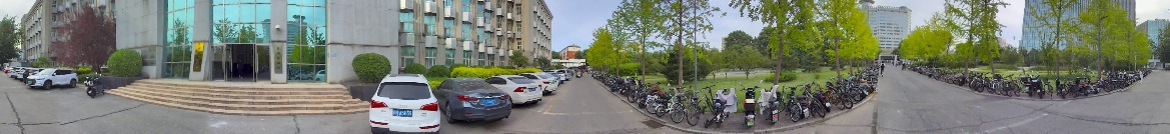}}

\caption{\textbf{TTWR’s Surround-view}: Illustration of the surround-view stitching results achieved for tractor-trailer wheeled robots (TTWRs) using our proposed method.}
\label{fig:final}
\end{figure}

%最终，我们在自己的数据集上实现了针对TTWRs的环视拼接，其结果如\ref{fig:final}所示。
Finally, we achieved surround-view stitching designed for TTWRs on our own dataset, with the results illustrated in Fig. \ref{fig:final}.

\section{Conclusion}
In this paper, we propose the Unified Vertex Motion Video Stabilization and Stitching framework for surround-view systems in tractor-trailer wheeled robots. Our approach navigates the complexities introduced by the non-rigid connections and diverse vibration patterns between the tractor and trailer, as well as the camera parallax challenges posed by the robot's large size. 

The framework's motion model has shown to simultaneously stabilize video and stitch images in real-time, providing robust performance in a variety of challenging scenarios. By employing a weighted cost function approach, our system adeptly balances the optimization of stitching and stabilization, yielding results that are superior to traditional static stitching methods, particularly in dynamic environments where the tractor is in motion. 

In our empirical tractor-trailer missions, we have substantiated the enhanced accuracy and practical utility of the proposed method, as well as its computational efficiency apt for real-time applications. We aspire to have provided a blueprint for advanced surround perception systems in large robots.

\section*{Acknowledgments}
 An earlier version of this paper was presented at the 2023 IEEE/RSJ International Conference on Intelligent Robots and Systems (IROS). The authors would like to thank Chunhui Zhu, Rundong Sun, Miaoxin Pan, Tao Wang, Kaixin Chen, Hao Han, Leyao Sun and all other members of the ININ Lab of the Beijing Institute of Technology for their contribution to this work.

\section{Declaration of Generative AI and AI-assisted technologies in the writing process}
During the preparation of this work the author used chatGPT in order to improve language and readability. After using this tool, the author reviewed and edited the content as needed and takes full responsibility for the content of the publication.

\bibliographystyle{elsarticle-num-names}
\bibliography{ref}


% %% For authoryear reference style

% \bibitem[Lamport(1994)]{lamport94}
%   \textit{\LaTeX: a document preparation system},
%   Addison Wesley, Massachusetts,

% \end{thebibliography}
\end{document}